%% file: neurips_2022.tex
\documentclass{article}
\PassOptionsToPackage{numbers}{natbib}

\usepackage[final]{neurips_2022}

\usepackage[utf8]{inputenc} 
\usepackage[T1]{fontenc}    
\usepackage[colorlinks=true,linkcolor=blue]{hyperref}       
\usepackage{url}            
\usepackage{booktabs}       
\usepackage{amsfonts}       
\usepackage{nicefrac}       
\usepackage{microtype}      
\usepackage{xcolor}         
\usepackage{graphicx}
\usepackage{graphics}
\usepackage{todonotes}
\usepackage{subcaption}
\usepackage{wrapfig}
\usepackage[most]{tcolorbox}
\usepackage{multicol}
\usepackage{enumitem}
\usepackage{listings}
\usepackage{pifont}
\usepackage[export]{adjustbox}

\include{macros}

\title{{\fontsize{15}{60}\selectfont Exploring Length Generalization in Large Language Models}}

\author{%
  Cem Anil~\thanks{Work done while interning at Google Research, Blueshift Team. Correspondence to: \texttt{anilcem@cs.toronto.edu, yuhuai@google.com, neyshabur@google.com}.}~~\textsuperscript{1, 3},~ Yuhuai Wu\textsuperscript{2}, Anders Andreassen\textsuperscript{1}, Aitor Lewkowycz\textsuperscript{1}\\
  \textbf{Vedant Misra\textsuperscript{1}, Vinay Ramasesh\textsuperscript{1}, Ambrose Slone\textsuperscript{1}, Guy Gur-Ari\textsuperscript{1}},  \\
  \textbf{Ethan Dyer\textsuperscript{1}, Behnam Neyshabur\textsuperscript{1}}\\
  \textsuperscript{1}~Google Research, Blueshift Team \\
  \textsuperscript{2}~Google Research \\
  \textsuperscript{3}~University of Toronto, Vector Institute
}

\begin{document}

\maketitle
\vspace{-0.1in}

\begin{abstract}
The ability to extrapolate from short problem instances to longer ones is an important form of out-of-distribution generalization in reasoning tasks, and is crucial when learning from datasets where longer problem instances are rare. These include theorem proving, solving quantitative mathematics problems, and reading/summarizing novels. In this paper, we run careful empirical studies exploring the length generalization capabilities of transformer-based language models. We first establish that naively finetuning transformers on length generalization tasks shows significant generalization deficiencies independent of model scale. We then show that combining pretrained large language models' in-context learning abilities with scratchpad prompting (asking the model to output solution steps before producing an answer) results in a dramatic improvement in length generalization. We run careful failure analyses on each of the learning modalities and identify common sources of mistakes that highlight opportunities in equipping language models with the ability to generalize to longer problems. 
\end{abstract}

\vspace{-0.1in}
\section{Introduction}

\begin{figure}[t]
\centering

    \includegraphics[width=0.82\textwidth]{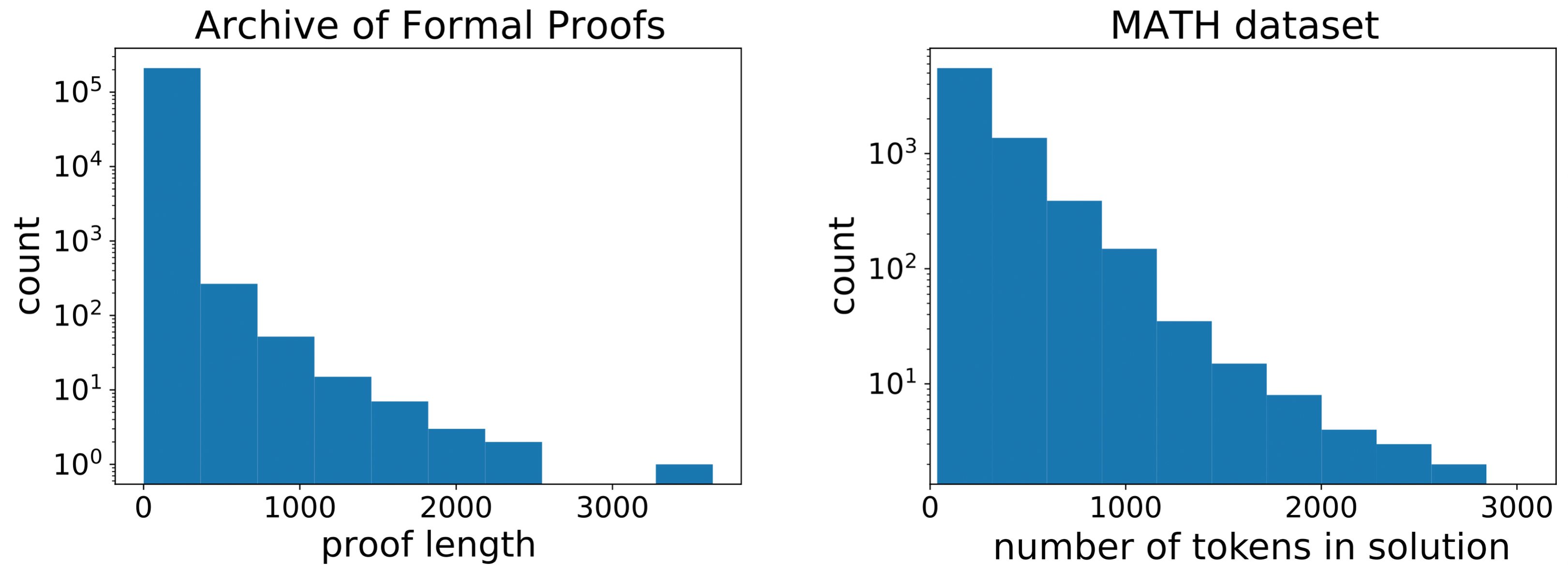}
    \caption{\small \textbf{Real world datasets have heavy tails in length: } \textbf{(left)} Histogram of lengths for proofs presented in the \href{https://www.isa-afp.org}{Archive of Formal Proofs} \textbf{(right)} Histogram of the number of tokens for solutions in the MATH dataset.~\citep{hendrycks2021measuring} }\label{fig:heavy_tail}
    \label{fig:second_hists}
    \vspace{-0.3cm}
\end{figure}

Many natural problems, such as theorem proving and program synthesis, have a notion of length that strongly correlates with the difficulty of the task. However, in these domains, the number of available problems typically drops rapidly as a function of problem length (e.g. Figure~\ref{fig:second_hists}). Hence, it is desirable to learn from examples of shorter lengths to generalize to longer ones or at least reduce the number of samples required for longer examples. We refer to this type of problem as \emph{length generalization}. 

Recent work on large language models (LLMs) has shown consistent improvement in their performance by scaling model and dataset size. However, such models are still incapable of length generalization. For example, \citep{gpt3} shows that even though scale helps with solving arithmetic problems, scale alone is likely insufficient for learning to solve instances of arbitrary lengths. This implies that models fail to learn the general algorithms that would enable this kind of generalization. Indeed, \citet{razeghi2022impact} showed that the performance of LLMs on mathematical calculations correlates with term frequency in the training data. This suggests that LLMs might have gained their current performance from surface-level memorization instead of learning to apply the correct algorithm. 

A recent line of work proposes to use a scratchpad, or chain-of-thought reasoning, when prompting LLMs~\citep{scratchpad, cot, lewkowycz2022solving} on multi-step tasks. Breaking down tasks into multiple small steps and presenting these steps to the model leads to improved performance across a variety of reasoning tasks including word problems, arithmetic, and code execution.

\begin{figure}
\centering
\includegraphics[width=0.78\linewidth]{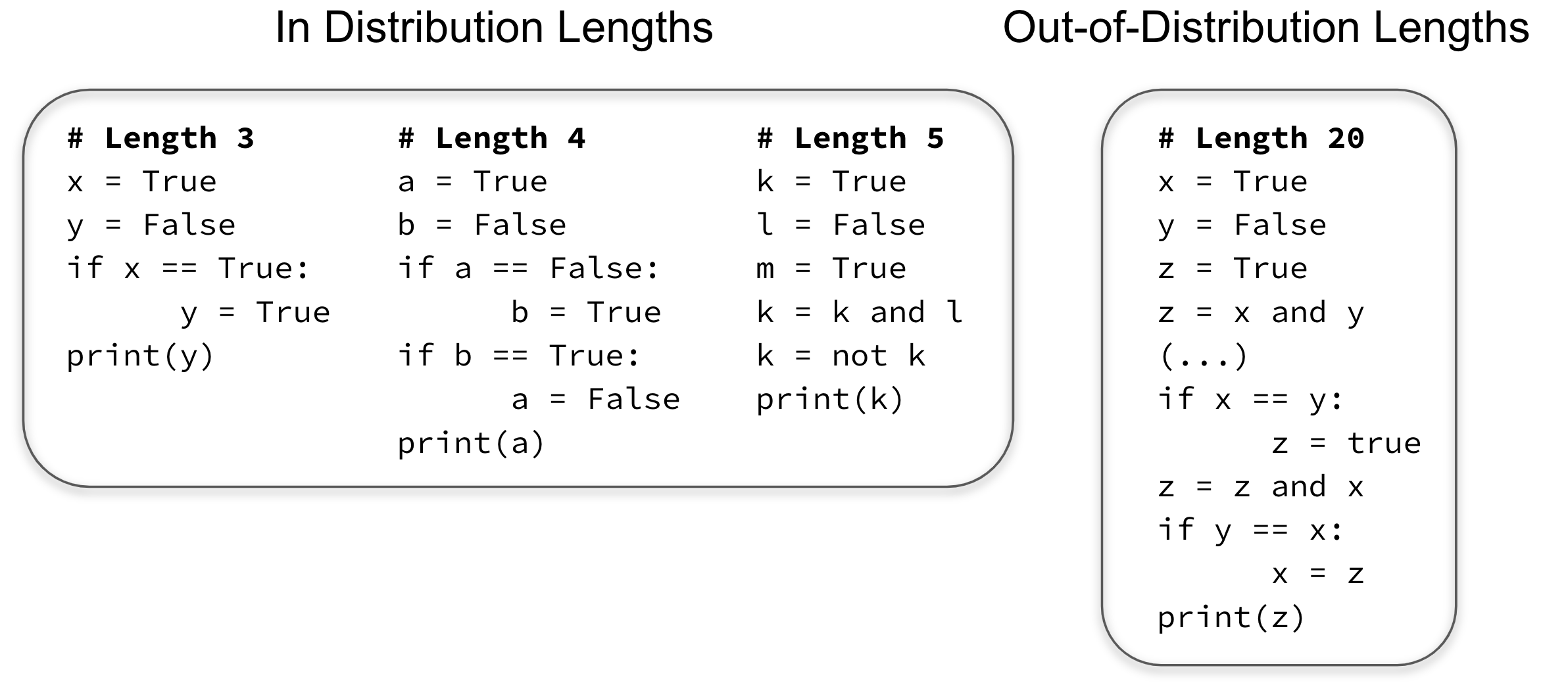}

    \caption{\small \textbf{Examples of variable assignment problems:} Can transformer language models learn from short instances of the Variable Assignment task (left) to extrapolate to much longer instances (right)? \textbf{Length generalization} is the ability to learn from shorter/easier instances of a problem to handle longer/harder instances. }
    \label{fig:var_assign_lg_example}

\end{figure}

We perform a systematic study of length generalization with transformer-based large language models. We consider problems in which learning an algorithm can in principle enable a model to extrapolate from short examples to problems of arbitrary length. In particular, we focus on two simple algorithmic tasks, \emph{parity} and \emph{variable assignment}, in which the model needs to keep track of a state in order to extrapolate to longer lengths (see Figure~\ref{fig:var_assign_lg_example}). These problems are illuminating because their simplicity allows us to probe the failure modes as well as contrast the learned solutions with the ground truth algorithm. They provide us with a setting to study how/when these large language models start to fail.

We study combinations of three kinds of techniques for LLMs: finetuning, few shot prompting (also referred to as in-context learning), 
and use of a scratchpad (also referred to as chain-of-thought), to understand the role of each method and the interplay among the three in length generalization. Interestingly, we observe non-trivial interactions among the three techniques; see Table~\ref{table:lg_options}.

\begin{table}[t]

\resizebox{\textwidth}{!}{%
\begin{tabular}{l|ccc}
\toprule
\textbf{Techniques}  & \textbf{In-distribution} & \textbf{Out-of-distribution} & \textbf{Improves with scale}  \\ \midrule
Fine-tune & \efficacyhigh & \efficacylow & \efficacylow \\
Prompting & \efficacylow & \efficacylow  & \efficacylow \\ 
Fine-tune + Prompting & \efficacyhigh & \efficacylow & \efficacylow \\
Fine-tune + Scratchpad & \efficacyhigh & \efficacylow & \efficacylow \\
Prompting + Scratchpad & \efficacymed & \efficacymed  & \efficacymed \\ 
Fine-tune + Prompting + Scratchpad & \efficacyhigh & \efficacyhigh$^{*}$ & \efficacyhigh \\
\bottomrule
\end{tabular}}
\vspace{0.1cm}
\caption{\small Performance on length generalization tasks of three techniques that language models admit: (1) Finetuning, (2) Prompting (or in-context few-shot learning) and (3) Scratchpad (Chain-of-Thought reasoning). We find that each technique (and the combinations thereof) have different modes of failure and present different trade-offs regarding in and out-of-distribution coverage. \efficacylow~signifies poor \efficacymed~signifies nontrivial, \efficacyhigh~signifies near-perfect performance. (*) Refers to task-dependency. }
\vspace{-0.5cm}
\label{table:lg_options}
\end{table}

\textbf{Contributions} Our main contributions are as follows:
\vspace{-2mm}
\begin{itemize}[leftmargin=5mm]
    \setlength\itemsep{0mm}
    \item We define and characterize the problem of length generalization using notions such as state tracking, execution depth, and per-step error rate. We study and carefully design two tasks, \emph{parity} and \emph{variable assignment}, that measure length generalization (Section~\ref{sec:lg}).
    \item We find that in the finetuning regime, scaling data, model sizes, and compute does not improve length generalization (Section~\ref{sec:scale}). We also observe that even when the model attains perfect in-distribution accuracy, it performs poorly in out-of-distribution domains. Surprisingly, different hyperparameter choices for finetuning have a large effect on length generalization performance, while having minimal effect on the final in-distribution performance (Section~\ref{sec:finetune_hyper}).
    \item We establish finetuning with scratchpad also fails to generalize to longer problems, in contrast to what is suggested by previous works~\citep{scratchpad}. We look into three potential failure cases: positional encoding, the presence of distractors, and end of token prediction, and conclude that distractors are the main culprit of failures for length generalization (Section~\ref{sec:scratchpad_finetuning}).
    \item We show that in the in-context learning regime, use of a scratchpad shows a qualitatively different behavior and significantly alleviates the decay of performance on longer problems. This capability is significant, as it implies that for LLMs, there are certain skills, like length generalization, that can be learned through in-context learning rather than through finetuning even in the presence of infinite data. This is in stark contrast to the common norms of machine learning (Section~\ref{sec:fewshot_finetuning}). 
\end{itemize}

\section{Length Generalization}
\label{sec:lg}

Many sequence tasks---especially ones that require reasoning capabilities---have problem instances that differ in terms of their \textit{lengths}. Shorter instances are often easier to state, process, and handle, and require less compute to find the answer. By contrast, longer instances are more challenging to parse and require more compute to solve. Tasks that have a \textit{reasoning component} are especially well represented in this category --- multi-hop reasoning~\citep{wang2019multi}, program execution~\citep{austin2021program}, deductive reasoning~\citep{clark2020transformers} and theorem proving~\citep{wu2020int}, to name a few. Note that having to deal with differing problem lengths poses two significant challenges. First, it is often the case that one encounters longer problem instances than the ones ever encountered during training, and is required to extrapolate. Second, even though longer problem instances have much more variety, real-world datasets often contain few long instances (see Figure \ref{fig:second_hists}). Both of these challenges are exacerbated if learning agents are not able to generalize across and beyond the lengths they learn from during training. This paper is about investigating to what extent transformer based language models are able to observe short problem instances and extrapolate to longer ones. 

\textbf{Instance Length as Number of Steps in a Markov Process}
It is possible to define \textit{problem length} in many different ways to capture different aspects of problem difficulty. Does there exist a notion of length that would expose the same length-generalization-related problem structure observed in qualitatively very different settings? Such a framing would enable researchers to design algorithms and interventions that have the potential to generalize across a broad range of tasks. To this end, we take the approach of characterizing length in the context of a \textit{deterministic Markov process}. From this perspective, length is simply the number of state transitions experienced by an initial world state. In other words, the data-generation process can be described as sampling an (1) \textit{initial state} and a (2) \textit{variable number of state transformations} to be applied sequentially on the initial state. The agent is provided both the initial state and the transformations, and is asked to predict the final state. This framing applies to a wide range of sequence problems, if not all of them---ranging from more mechanical tasks such as code and algorithm execution and theorem proving, to less structured tasks, such as solving math problems and summarizing novels.\

In our empirical investigation we focus on two synthetic tasks: \textit{parity} and \textit{variable assignment}. These tasks avoid problem-specific subtleties that could mislead our analyses, while strongly capturing the deterministic Markov process structure.  

\subsection{Tasks}

\textbf{Parity:} The parity task is an age-old learning problem that requires the trained agent to predict whether a bit-string has an even or odd number of ones in it. For example, the parity of the bitstring $[0, 1, 1, 0, 1]$ is ``odd" (or $1$) as opposed to ``even" (or $0$), because there is an odd number of $1$s in the bit-string. The parity task admits a sequential solution that enables length generalization in a straightforward way: simply process the bits left-to-right and record the parity of the bits processed so far as the state. The default notion of length in the parity task is the number of bits in the input. However, we also experiment with a version where the number of bits is kept constant, and the number of $1$s (i.e. the parity flipping bit) is systematically varied. The number of $1$s stands for the number of state changes contained in the input bit-string, and actually appears to capture a more relevant notion of length for transformer models (see Section \ref{sec:scale}).

\textbf{Boolean Variable Assignment Task: } The Boolean Variable Assignment task is designed to capture arbitrarily long, potentially branching unidirectional execution flows. An instance of this task can be seen in Figure \ref{fig:var_assign_lg_example}. The inputs consist of semantically correct (i.e. bug-free) Python programs in which each line contains a boolean variable assignment operation. The output is simply the value of the variable presented in the final line of the program. The sequential solution to this task is to simply execute the program line by line while keeping track of the state of all variables.

\begin{figure}
\centering
\begin{subfigure}{0.45\textwidth}
    \includegraphics[width=\textwidth]{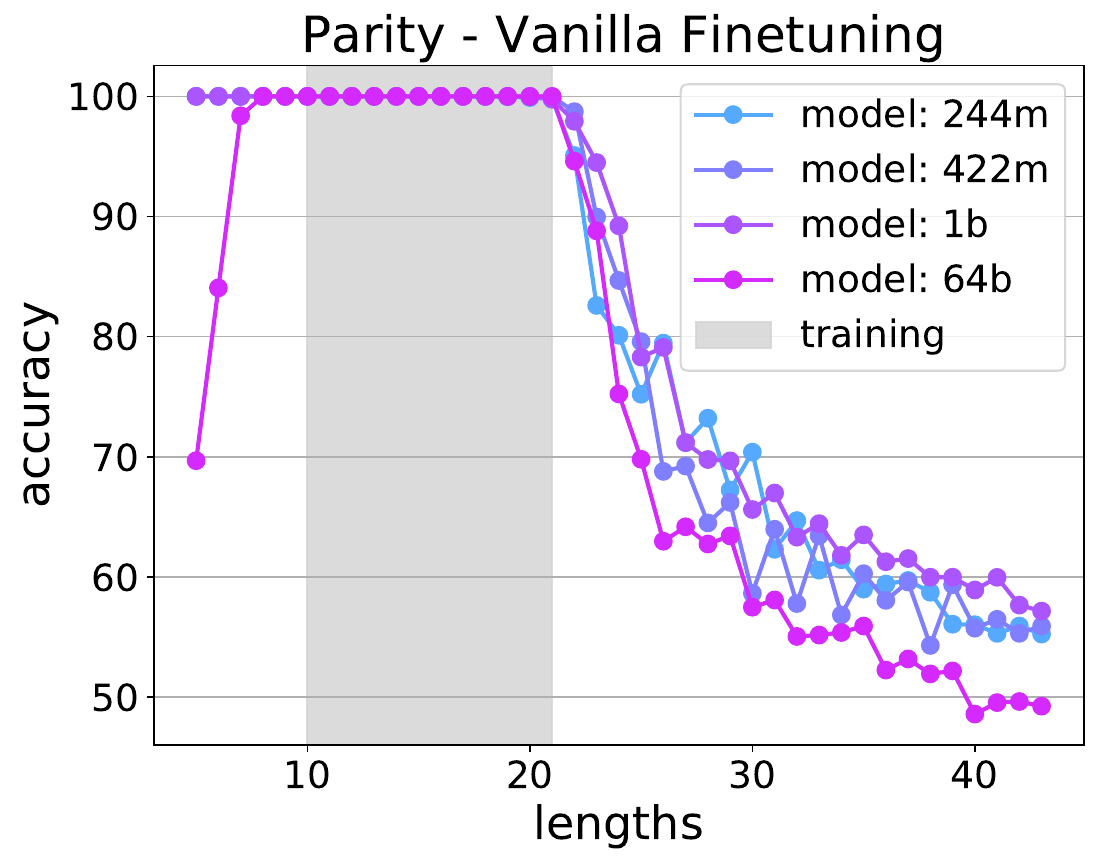}
\end{subfigure}
\hfill 
\begin{subfigure}{0.45\textwidth}
    \includegraphics[width=\textwidth]{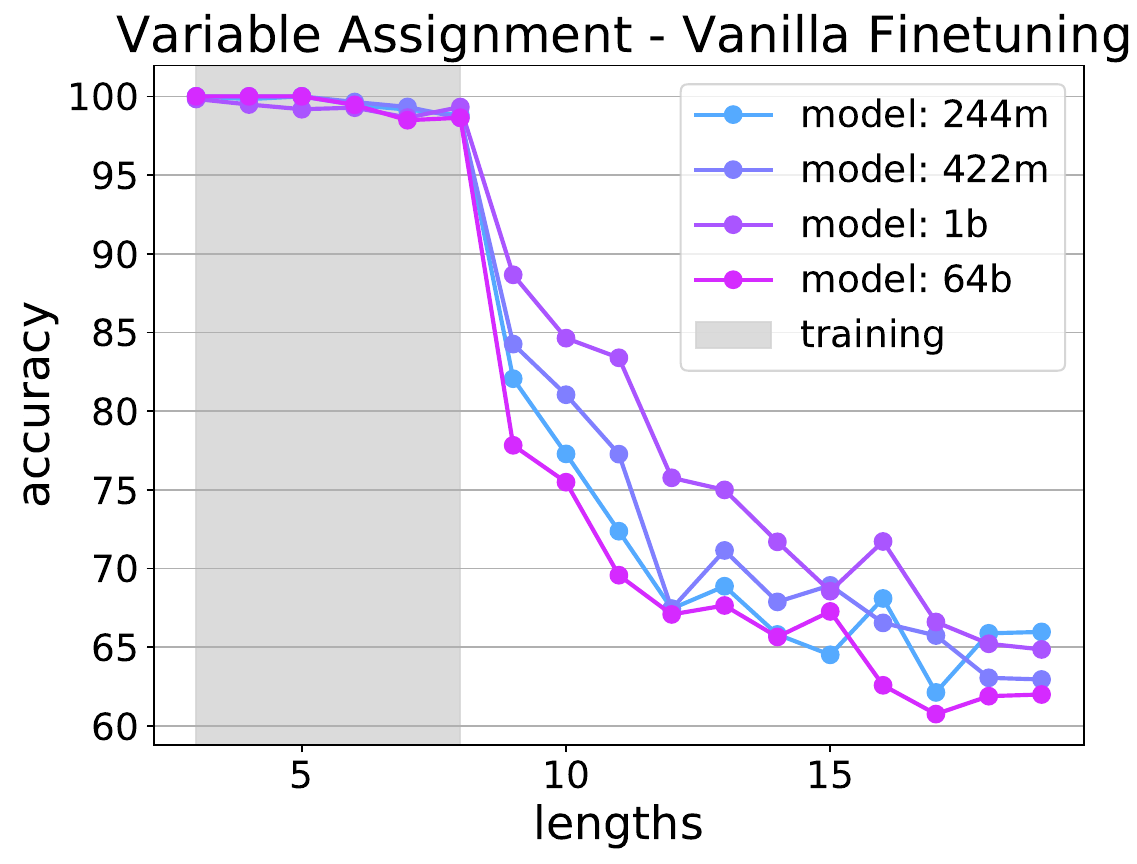}
\end{subfigure}
\vspace{-3mm}
    \caption{\small \textbf{Finetuned Length Generalization performance doesn't improve with scale:} Models of vastly different scales fail at length generalization on both Parity and Variable Assignment tasks, displaying identical generalization pathologies. The x-axis represents problem length and the y-axis represents the accuracy attained at that problem length. The training lengths are highlighted in grey. }\label{fig:no_scratchpad_lg}
\vspace{-4mm}
\end{figure}

The data generation procedure involves randomly generating execution flows that involve Boolean operations;  see Supplementary Material (SM) for details.

We focus our evaluations on two variants of this dataset. (1) The \textit{diverse variable assignment} split consists of a wide range of boolean operators available and is intended to contain maximally diverse programs.  (2) The \textit{chain-like variable assignment split} consists only of operations that compose the values of already defined variables. This results in long chains of dependencies between the initial values of the variables and the queried one, ensuring that there are almost no redundant operations in the program (i.e. operations that can be removed without affecting the output of the program). This split emphasizes the sequential nature of the variable assignment problem. 


\vspace{-0.1in}
\section{Standard Finetuning Fails at Length Generalization}
We begin by demonstrating that finetuning transformer models on length-generalization tasks results in poor out-of-distribution performance. In experiments we use \pretrainedmodel \footnote{We omit the name of the model for the purpose of anonymization.} decoder-only models. 
These checkpoints were trained using general natural language data. We use the AdaFactor optimizer~\citep{shazeer2018adafactor} during finetuning, and tune the learning rate, batch size and dropout. We trained the networks until the in-distribution validation accuracy settles (20000 gradient steps for parity and 18000 gradient steps for variable assignment). The loss was only computed on the target tokens (i.e. the model wasn't trained to model the input questions). 

\subsection{Scale Doesn't Improve Length Generalization}
\label{sec:scale}
\textbf{Parity:} We finetuned four pretrained \pretrainedmodel models with 244m, 422m, 1b and 64b parameters on the parity task, where the training distribution included randomly sampled bitstrings of length 10 to 21. We then evaluated the performance on bitstrings of length 3 to 40; see Figure \ref{fig:no_scratchpad_lg}. We find that model scale has a little effect on length generalization. 

\begin{figure}
\centering
\begin{subfigure}{0.45\textwidth}
    \includegraphics[width=\textwidth]{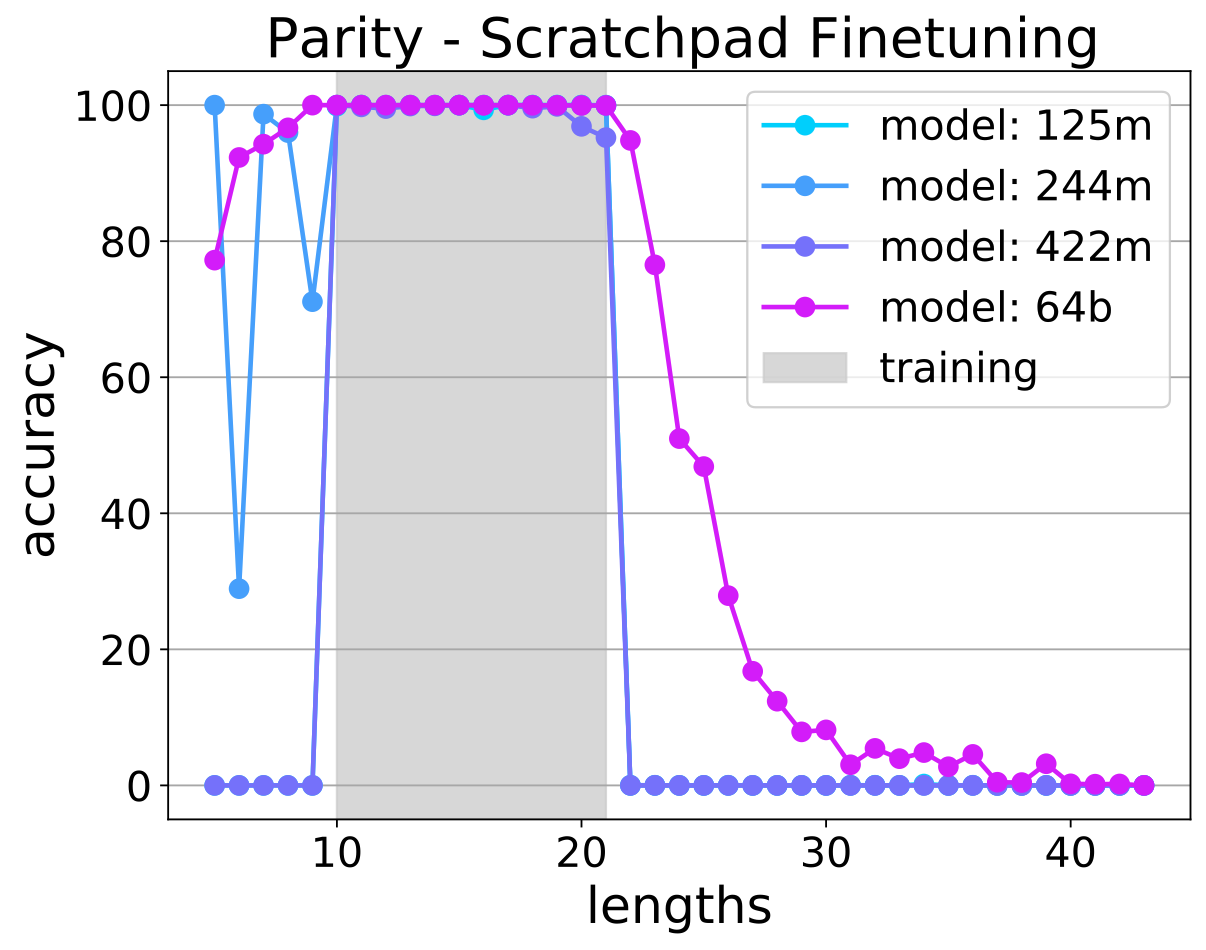}
\end{subfigure}
\hfill
\begin{subfigure}{0.45\textwidth}
    \includegraphics[width=\textwidth]{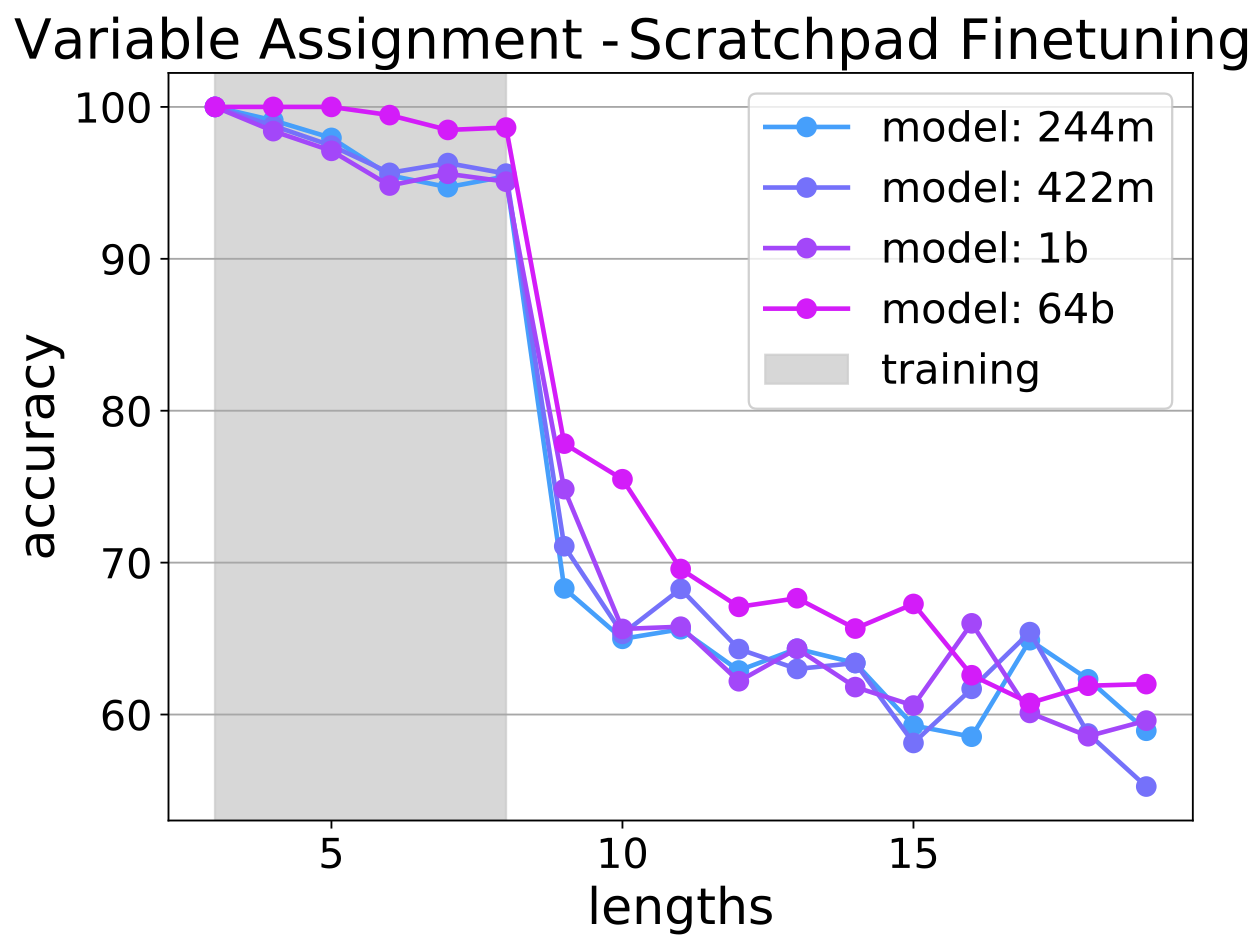}
\end{subfigure}
\vspace{-3mm}
\caption{\small \textbf{Scratchpad finetuning displays poor length generalization:} Scratchpad finetuning displays qualitatively similar length generalization pathologies as vanilla finetuning. The x-axis represents problem length and the y-axis represents the accuracy attained at that problem length. The training lengths are highlighted in grey.}
\label{fig:scratchpad_lg}
\vspace{-3mm}
\end{figure}

\textbf{Variable Assignment:} We finetuned the same models
on the \textit{chain-like} Variable Assignment Task, described in Section \ref{sec:lg}.  We kept the in-distribution lengths at 3 to 8, and evaluated the test performance on lengths 3 to 19. The results can be seen in Figure \ref{fig:no_scratchpad_lg}. Just like in the parity task, while the in-distribution performance is (near) perfect, out-of-distribution performance degrades rapidly as length increases. To get a sense of just how weak the out-of-distribution performance is, we also trained a 422m model on the same dataset, except we shuffled the operations before feeding it to the model. This removes the sequential dependency between the operations, and helps us establish a strong baseline that only predicts the answers based on non-sequential, spurious correlations. The accuracy-length curves for the baseline can be found in SM. 

\vspace{-2mm}
\subsection{Transformers Prefer Parallel Strategies over Sequential Ones} 
\label{sec:finetune_failure}
The results presented in Section~\ref{sec:scale} establish that, when presented with sequential length generalization problems, transformers are biased toward learning non-sequential ``shortcut'' solutions that fail at longer problem instances. We ran additional experiments to gain a better understanding of the nature of this generalization pattern.

\begin{figure}
    \centering
    \begin{subfigure}{0.4\textwidth}
        \includegraphics[width=\linewidth]{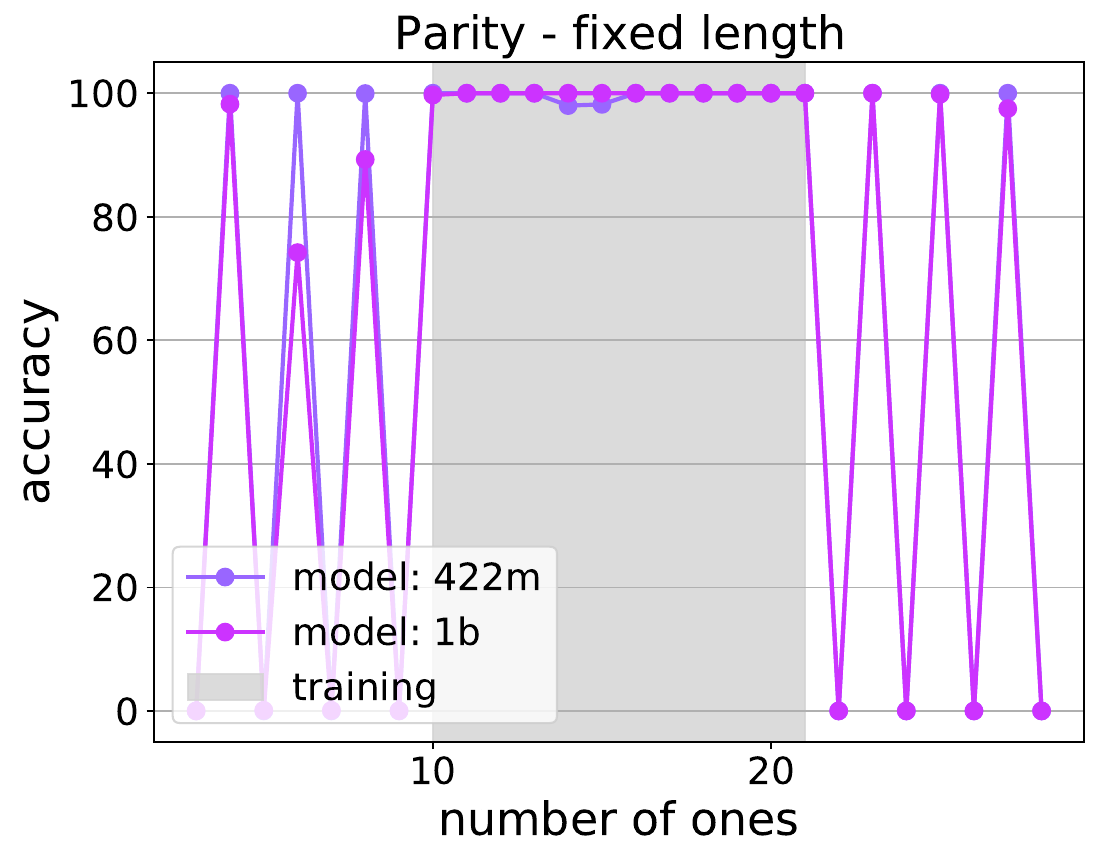}
          \label{fig:parity_vary_num_ones}
    \end{subfigure}
    \hfill
    \begin{subfigure}{0.58\textwidth}
        \includegraphics[width=\linewidth]{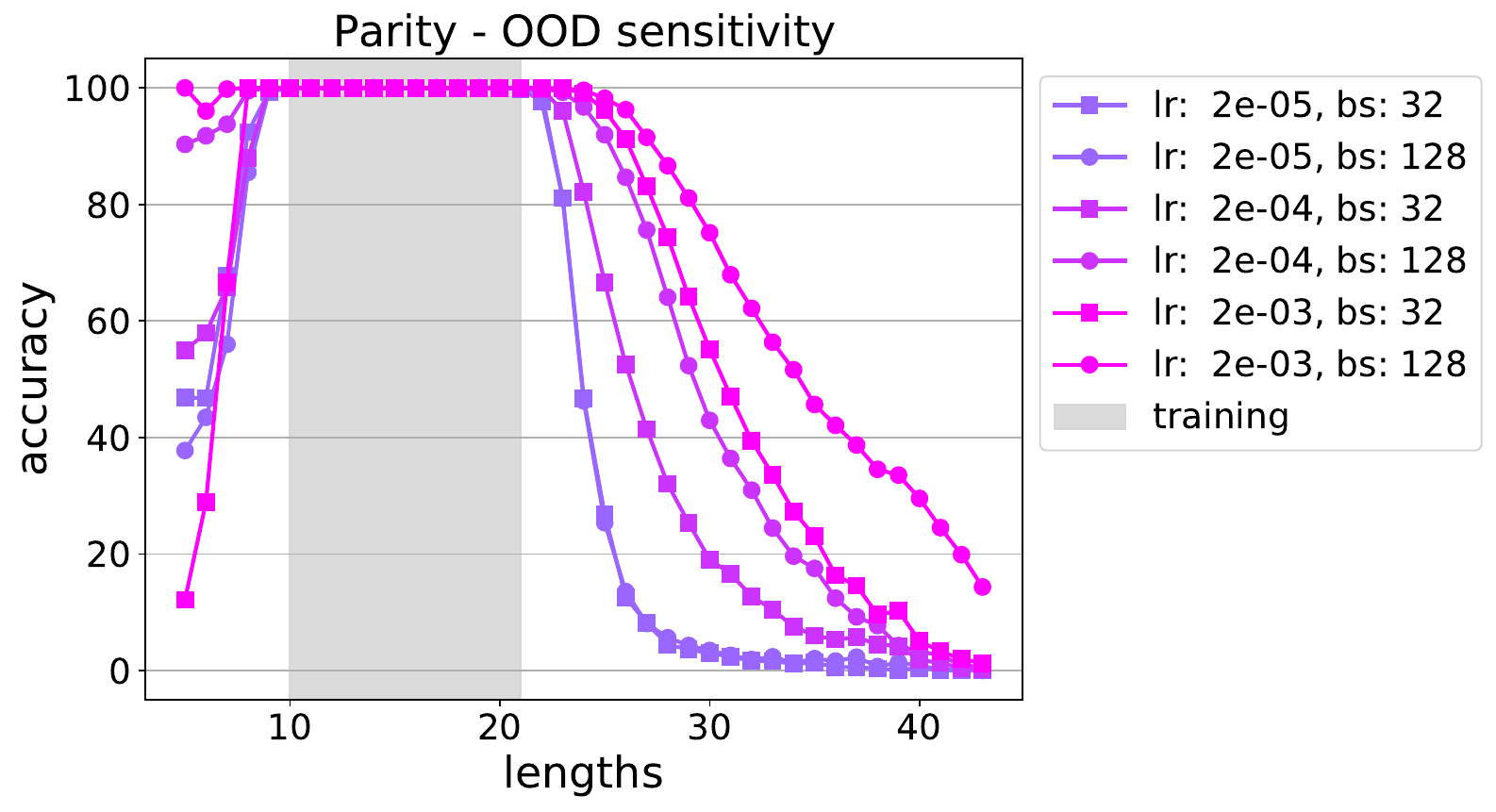}
        \label{fig:finetune_hyper}
    \end{subfigure}
    \vspace{-3mm}
    \caption{\small \textbf{(left) Complete lack of length generalization:} Transformers trained on the parity task have difficulty generalizing to bit-strings that have a different number of $1s$. \textbf{(right) Sensitivity to hyperparameters:} Trained networks sharing architecture, data and in-distribution loss can have very different length generalization performances. \textit{lr} stands ``learning rate" and \textit{bs} stands for ``batch size". }
    \vspace{-2mm}
    \label{fig:lg-subtleties}
\vspace{-3mm}
\end{figure}

On parity, we ran finetuning on a different distribution of bit-strings: Instead of first randomly sampling the number of bits in the input bit-string, then sampling the values of the bits, we fixed the total number of bits in the input, and only varied the \textit{number of ones} in the bit-string uniformly. We trained with $10$ to $20$ ones in the input distribution and tested on an interval containing $1$ to $30$ ones. This makes sure that the number of tokens (now fixed at $30$) is now disambiguated from number of state changes, which for parity is equal to the number of ones. The difference between in and out-of-distribution performance is even starker for this data distribution (Figure \ref{fig:lg-subtleties}): while in-distribution performance was $100$\%, OOD performance was roughly equivalent to random prediction\footnote{The periodic $0$\% and near $100$\% performance on OOD lengths is due to the models' tendency to output $0$ or $1$ depending on whether the input has a significantly higher ratio of $0$s or $1$s. On average, the accuracy on OOD length is not better than random guess. }. This suggests that the transformers are learning a non-sequential solution that involves counting the number of ones in the input, and then thresholding the output. This is not surprising, given that self-attention is an equivariant transformation capable of performing pooling operations like max-pooling~\citep{lee2019set}. This strategy doesn't allow for knowledge transfer between problems of different lengths. Note that this bottom-up counting behaviour is complementary to the left-to-right counting behaviour displayed by recurrent models~\citet{suzgun2019lstm}. 
On the variable assignment dataset, we finetuned a 255m \pretrainedmodel model on the \textit{diverse} split of the variable assignment dataset of programs up to 16 lines, and evaluated on the same data generating distribution up to 32 lines. We measured the evolution of the model's accuracy with respect to training iterations on different program lengths (quantified by number of lines). The results are in SM.

We again observed that a different notion of length (which we call \textit{computational graph depth}) captures the difficulty of problem instances better than number of program operations. A variable assignment program can be represented as a computational graph where each node corresponds to a variable, and each edge corresponds to an operation. Computational graph depth is the length of the longest dependency chain that connects to the queried variable node. This notion of length corresponds to the highly parallelizable strategy of executing programs by iteratively resolving computational graph dependencies. We present two results that suggest that computational graph depth is a more relevant notion of length for transformers. (1) Inspecting the order of problem instances in which the trained transformer correctly solves this task, we find that performance is strongest on examples with small computational graph depth, even if these examples are long in terms of number of operations. (2) The transformer does a good job of handling programs with an out-of-distribution number of operations, but for which computational graph depth is in-distribution. 

\subsection{In-Distribution Generalization Doesn't Predict OOD Generalization on Length Generalization Tasks}
\label{sec:finetune_hyper}

Prior work on out-of-distribution generalization establishes that in many tasks, in-distribution loss is a strong predictor of out-of-distribution generalization~\citep{nagarajan2020understanding}. Our experiments on the parity task indicate that the distribution shift induced by changing problem lengths falls outside of the this category. Figure~\ref{fig:lg-subtleties} shows how the same model trained on the same data achieving roughly the same in-distribution cross entropy loss behaves on OOD data, where the difference is solely induced by the choice of different hyperparameters.

\vspace{-0.1in}
\section{Scratchpad Finetuning Still Fails at Length Generalization}
\label{sec:scratchpad_finetuning}

It has been shown in prior work that it's possible to get pretrained LLMs to solve a given task by not only outputting the answer, but also the solution steps behind it. \citet{nye2021show} use scratchpad finetuning to achieve strong in-distribution performance on execution based tasks such as code execution and computing polynomials. While they also  report modest length generalization results on integer arithmetic, we find that scratchpad finetuning suffers from similar length generalization pathologies than vanilla finetuning does.  The results on parity and variable assignment tasks can be seen in Figure~\ref{fig:scratchpad_lg}. The precise scratchpad strategies used for these tasks are described in detail in SM.

\textbf{Error analysis: } To understand the causes of failure in training scratchpad strategies, we focused on two architectural choices that could account for the poor performance: (1) how transformers encode position information, and (2) whether the transformers are trained to predict an end-of-sequence (EOS) token. \pretrainedmodel models use T5 position biases \citep{raffel2019exploring} to handle position information. If the network is only trained with short instances, position biases that handle longer positional distances might not be trained, explaining poor length generalization. Similarly, \citet{newman2020eos} report that networks trained with EOS token prediction often suffer from generalizing to longer problem instances, because of the models' tendency to emit EOS tokens prematurely, as well as the EOS tokens' effect on the representations that get learned. 

We tested the extent to which these effects can explain lack of length generalization as follows. We padded both the input bit-strings and the scratchpad content with dummy padding tokens to make the token count the same. We also augmented the input and scratchpad targets with the same number of padding tokens on the left and right so that the relevant bit to attend to when executing the sequential scratchpad strategy corresponds to the same T5 position bias bin. Examples of the updated input-target pairs can be seen in SM. While this intervention helps, the trained models still display significant length generalization issues. 

\begin{figure}
    \centering
    \begin{subfigure}{0.49\textwidth}
    \includegraphics[width=\textwidth]{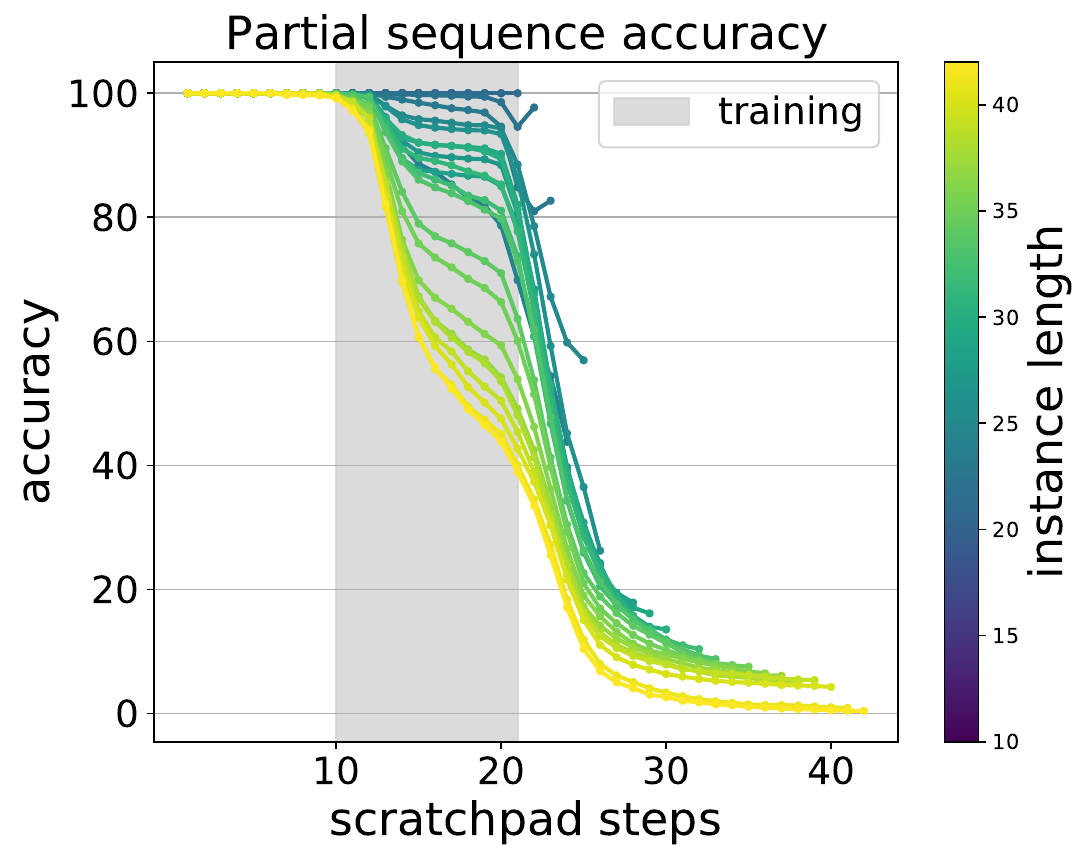}
        \label{fig:first_k_tokens}
    \end{subfigure}
    \hfill
    \begin{subfigure}{0.49\textwidth}
        \includegraphics[width=\linewidth]{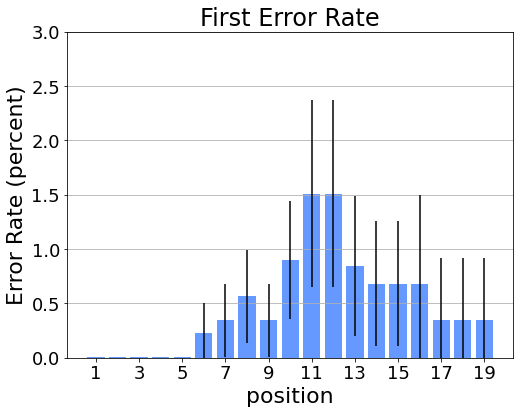}
        \label{fig:exa}
    \end{subfigure}
    \caption{\small \textbf{(left) Effect of input length on per-step scratchpad accuracy:} Points corresponds to the accuracy (y-axis) of the first $x$ scratchpad steps (x-axis) on parity instances of variable length (color). If the input length is out-of-distribution, even in-distribution scratchpad steps are inaccurate, implying the model hasn't learned an attention pattern that generalizes to longer bit-strings. \textbf{(right) Roughly constant per-step error rate:} The per-step error rates of the \pretrainedmodel 128b model, few-shot finetuned on the coin-flip version of the parity task remain roughly constant across the scratchpad steps. This is in stark contrast with zero-shot scratchpad finetuned models, where the per-step error rates increase abruptly when the model is evaluated on OOD lengths. }
    \label{sec:position_errors}
\vspace{-5mm}
\end{figure}

To gain further insight about the source of the problem, we plotted how the scratchpad target prediction error rates change as a function of (1) how far along one is in constructing the scratchpad, and (2) the length of the input bit-string. The results can be seen in Figure \ref{sec:position_errors}. The fact that the model makes mistakes in in-distribution scratchpad steps when the input has an OOD length implies that the attention mechanism isn't capturing the relevant part of the input to form the scratchpad output. 
See SM for additional analysis.

\vspace{-0.1in}
\section{Scratchpad Prompting Significantly Improves Length Generalization}
\label{sec:fewshot_finetuning}
\citet{cot}, \citet{nye2021show} and \citet{lewkowycz2022solving} showed that combining prompting (i.e. in-context learning) with scratchpad strategies present a powerful combination. They demonstrate that pretrained LLMs, without the help of any finetuning, can solve grade school math word problems and execute pieces of code with nontrivial correctness~\citep{nye2021show}, when prompted with the right scratchpad strategy. We corroborate these findings, and report that scratchpad prompting endows pretrained LLMs with the capability of \textit{variable length template matching} (see Figure~\ref{fig:few_shot_lg}). That is, in-context learning enables the model to ``learn" solution steps from a small number of short instances, and apply the same template on significantly longer instances with a high degree of accuracy. 

\vspace{-2mm}
\subsection{Few-shot scratchpad} 
\label{sec:few-shot-sp}
Contrary to vanilla and scratchpad finetuning, we find that under the right conditions, few-shot scratchpad strategies sometimes significantly improves LLMs' capability to extrapolate to lengths much further than what pretraining weights grant them.

\begin{figure}
\centering
    \includegraphics[width=0.75
    \textwidth]{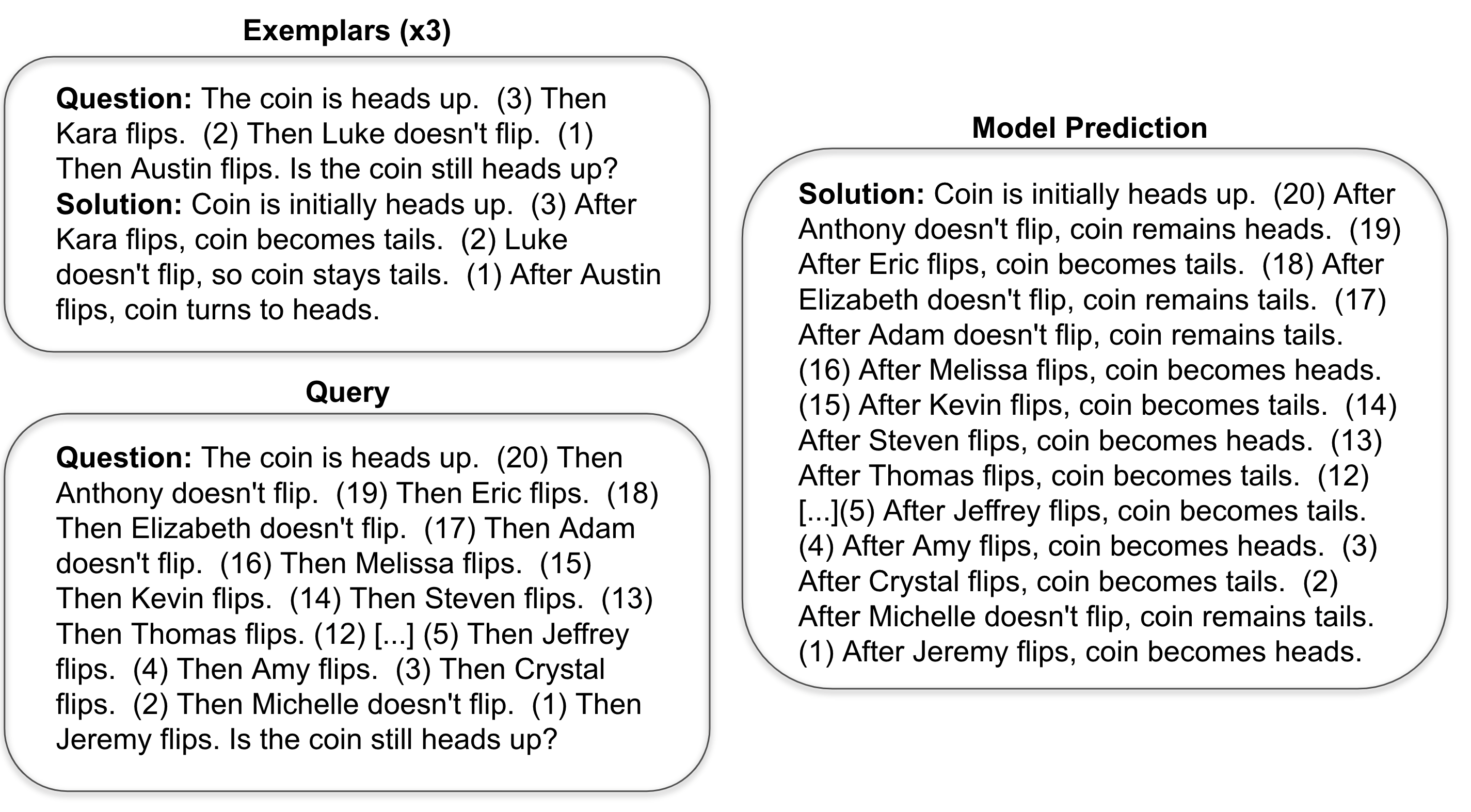}

    \caption{\small
    \textbf{Few-shot length generalization:} The largest \pretrainedmodel model is able to map the scratchpad solution template from a few short exemplars onto much longer queries. }
    \label{fig:few_shot_lg}
\vspace{-3mm}
\end{figure}

To evaluate the performance of few-shot conditioning with scratchpad inputs without any finetuning, we phrase the parity problem in natural language as a coin flipping task. An example for the few-shot prompts we used can be seen in Figure \ref{fig:few_shot_lg}. \citet{cot} also report results on the coin-flip task: the scratchpad format we used differs from theirs in that while ours respects the sequential nature of the task (i.e. each coin flip corresponds to a step in the scratchpad solution), \citet{cot}'s scratchpad strategy involves summing up the number of coin flips, then deciding on the final output based on the evenness/oddness of the sum. Also, while they only test up to $4$ flips, we go up to $20$ flips while still attaining highly nontrivial accuracy levels. 

For the variable assignment task, our scratchpad strategy involves copying over the program that's being executed, with comments added in between lines specifying the value of the variable that was assigned in the line above. Instances of this scratchpad strategy can be seen in SM. 

\begin{figure}
\centering
    \includegraphics[width=0.48\textwidth]{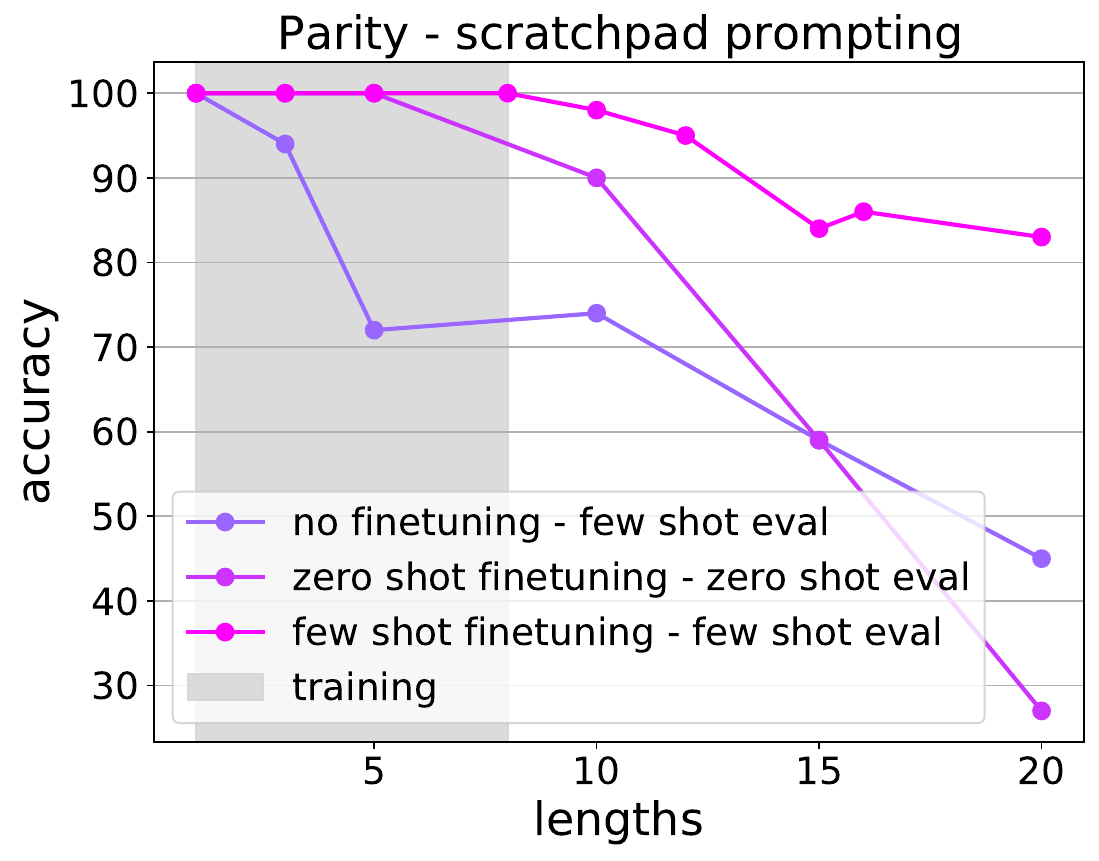}
    \includegraphics[width=0.49\textwidth]{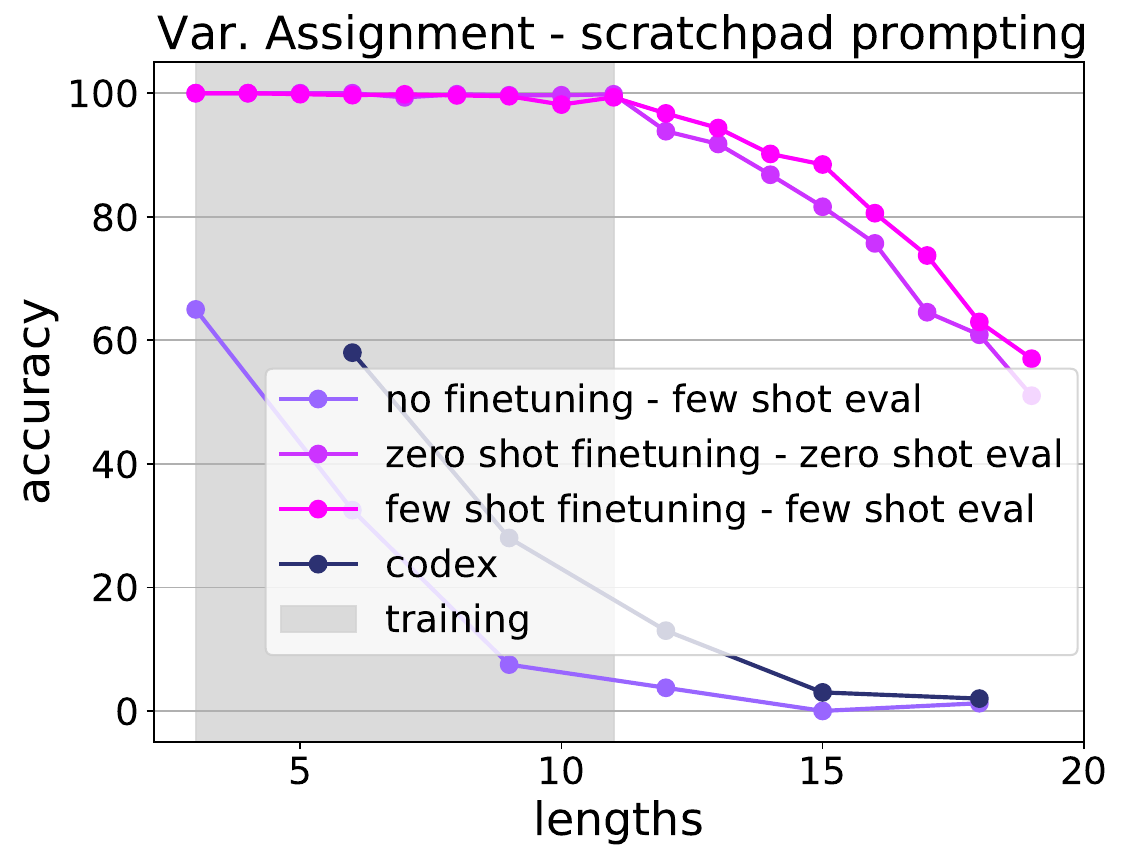}
\caption{\small \textbf{Few-shot finetuning with scratchpad} displays qualitatively different behaviour on parity and variable assignment tasks. On parity, where the non-finetuned model already performs very well, few-shot-finetuning with scratchpad leads to a significant performance boost over zero-shot finetuning with scratchpad. On variable assignment, where the base model doesn't perform poorly, there's not a significant gap between few-shot finetuning and zero-shot finetuning with scrathpad. The performance of OpenAI's Codex model \citep{chen2021evaluating} on the variable assignment task is also provided. }
\label{fig:fewshot_scratchpad_lg}
\vspace{-5mm}
\end{figure}

Figure \ref{fig:fewshot_scratchpad_lg} shows the performance 
of the pretrained \pretrainedmodel 128b model on the coin-flip version of the parity task. Figure \ref{fig:few_shot_lg} shows an instance of how a 
length 3 prompt
can induce the model to correctly output a $20$ step scratchpad. We find that with the right scratchpad prompt, 
LLMs are able to generate correct scratchpad solutions.
This reduces the problem to simply filling in the content of the generation correctly by inferring the right state transitions without having to figure out how to extrapolate the solution template.

\textbf{Few-Shot Finetuning with Scratchpad Strategies: } 
Does combining finetuning, few-shot prompting, and scratchpad strategies improve length generalization?

We find that the answer is \textbf{yes} in the case of parity. As seen in Figure \ref{fig:fewshot_scratchpad_lg}, few-shot finetuning performs significantly better than the baseline model, both on in- and out-of-distribution lengths. Note that the vanilla (i.e. no shot) finetuning baseline also outperforms the no-finetuning baseline, it actually does worse on the larger lengths --- a pathology that doesn't appear with few-shot finetuning. 

The results point to a qualitatively different picture for the variable assignment task. Both few-shot finetuning and vanilla finetuning result in similar length generalization behavior (Figure \ref{fig:fewshot_scratchpad_lg}). We hypothesize that this distinction is caused by the different pretrained performances that the model displays on these tasks: while length generalization is already strong with no finetuning on parity, that's not the case for variable assignment. In the latter case, the model is forced to acquire a new skill via finetuning, which displays the same pathologies as zero-shot finetuning with scratchpad. As a sanity check, we evaluated the (few-shot) finetuned performance of the pretrained model on an alternative, synthetic prompt style that yields poor performance without any pretraining: As expected by the aforementioned hypothesis, we observed that the few-shot finetuned model on this task also shows significant length generalization pathologies. The results can be found in SM. We leave a more rigorous evaluation of this hypothesis as future work.

\vspace{-3mm}
\section{Related Works}
\vspace{-1mm}

There have been many attempts to study generalization from shorter/easier to longer/harder examples. 

\textbf{Challenges in length generalization: } Several existing works have investigated pathologies that arise when models are asked to generalize to processing and generating longer (measured by number of tokens) sequences. \citet{newman2020eos} find that sequence models trained with and in the absence of the end-of-sequence token display qualitatively different length extrapolation behaviour and learn different representations. \citet{dubois2019location} proposes modifications to the commonly used dot-product attention to improve the models' ability to extrapolate to longer sequences. \citet{murray2018correcting} demonstrate that neural machine translation models tend to have a bias towards generating shorter-than-desired translations. \citet{yehudai2021local} show that length generalization issues are also present in training graph neural networks, where extrapolating across graph size presents a challenge. \citet{ju2021staircase} propose a new attention mechanism to facilitate recurrent processing in transformer models. \citet{alibi} propose modifying transformer attention biases to facilitate generalization beyond the training context length. Concurrent work \citep{zhang2022unveiling} propose a synthetic dataset named LEGO (Learning Equality and Group Operations), an instantiation of which resembles our variable assignment task where the only boolean operations allowed are \textit{assign} and \textit{negate and assign}, and overriding the values of variables is not allowed. Their analyses on OOD generalization largely complement ours: while we focus on decoder-only architectures and scratchpad strategies as a way of carrying over state, they focus on encoder-only architectures, and investigate the effect of weight-sharing. 

\textbf{Easy-to-Hard generalization:} \citet{schwarzschild2021can} and \citet{bansal2022end} use weight-tied neural networks to generalize from easy to hard examples. \citet{schwarzschild2021can} also provide three tasks to benchmark easy-to-hard generalization.
\citet{dehghani2018universal} and \citet{kaiser2015neural} assess the capabilities of their proposed architectures on easy-to-hard generalization problems. 

\textbf{Inductive Biases Related to Lenght Generalization:} \citet{mccoy2020does} study the inductive bias of seq-to-seq learners on English question formation and English tense reinflection tasks and find that LSTM and GRU networks often display differing strategies, caused by the use of differing activation functions. \citet{suzgun2019lstm} find that recurrent networks can perform dynamical counting, and encode hierarchical representations, which enables them to solve nontrivial Dyck tasks using k-counters. \citet{kharitonov2020they} also study the inductive bias of different architectures, and conclude that transformer and LSTM architectural have a tendency to learn hierarchical strategies, whereas CNN based strategies display more linear structure. \citet{he2019unlearn} propose a method to learn natural inference models that are not biased on spurious correlations. \citet{mccoy2019right} show that transformer models that display strong performance in natural language inference can have superficial biases that fool them in systematic ways and proposes a framework to think about these biases. 

\vspace{-3mm}
\section{Conclusion}
\vspace{-1mm}
The ability to learn from shorter/easier problem instances to generalize to longer/harder ones is a key capability in a large number of tasks, especially ones requiring reasoning. We defined the concept of \textit{length generalization} and measured language models' length generalization capabilities. After conducting careful experiments using finetuning, scratchpads, and few-shot prompting, we reached the following conclusions: (1) Generalizing in length is a challenge for language models at least up to the 100B parameter scale.
Both vanilla finetuning and finetuning with scratchpads suffer from a lack of length generalization caused by models' tendency to pick up non-sequential pattern that don't apply to longer problem instances. (2) Few-shot scratchpad prompting enables pretrained large language models to pick up scratchpad-templates that extrapolate to arbitrary lengths, leading to dramatic improvements on longer problem instances. Unlike raw finetuning, this approach does scale with model size~\cite{cot}.  (3) Trying to further enhance the performance of few-shot scratchpad prompted LLMs via finetuning yields mixed results, depending on the non-finetuned performance of the base model at the target task. We emphasize that the aforementioned few-shot variable length pattern matching capability --- something that doesn't require changing model architecture ---  offers a qualitatively different approach to handle length generalization in contrast to prior art that introduced architectural modifications to achieve the same goal. This capability is also significant in that it implies that for LLMs, there are certain skills, like length generalization, that can be learned better through in-context learning rather than through finetuning, even in the presence of infinite data.

\bibliographystyle{unsrtnat}
\bibliography{references}

\newpage
\appendix

\section{Data Generation Details}
We describe the data generation procedures for the parity and variable assignment tasks in detail. 

\subsection{Parity Datasets:} 

\begin{figure}
\centering
    \includegraphics[width=0.80
    \textwidth]{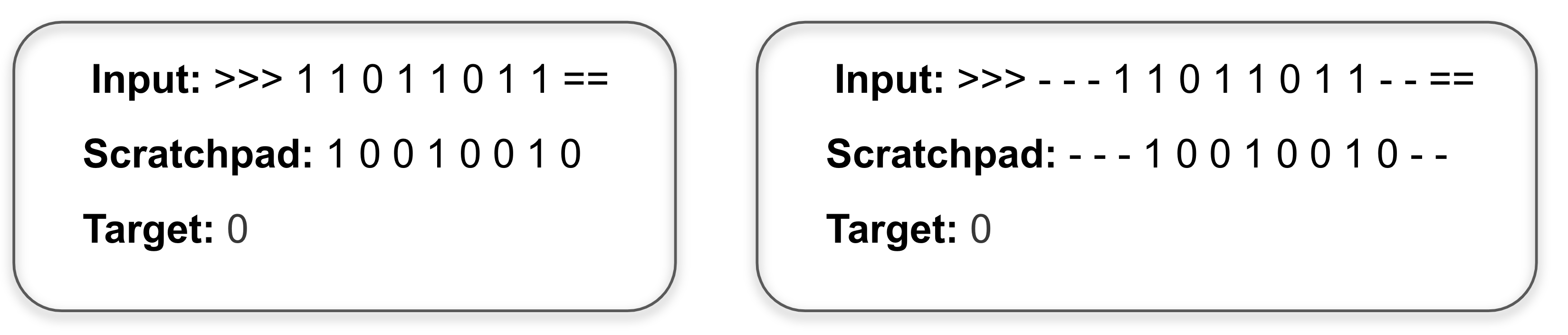}

    \caption{\small
    \textbf{Parity problem instances:} \textbf{(left)} A sample $8$-bit parity problem instance, along with the scratchpad targets. The scratchpad represent the intermediate parity state as the sequence is processed left to right. \textbf{(right)} A parity problem instance with the padded scratchpad strategy. Both the input and the scratchpad targets are padded left and right with the same number of padding tokens, such that the relevant bit to attend to while constructing the scratchpad bits is always equidistant even when there are different number of bits in the input. }
    \label{fig:parity_instances}
\end{figure}

\textbf{Synthetic Parity Dataset:} A $8$-bit example of the synthetic parity example, along with the corresponding scratchpad targets, can be seen in Figure \ref{fig:parity_instances}. We added the prefix ``> > >" to signify the start of the parity sequence, and the suffix ``==" to signify the start of the target or scratchpad tokens. There's no special meaning associated with the particular prefixes and suffixed used. 

We experimented with two version of the synthetic parity dataset: In one split, we varied the \textit{number of bits} in the input, and in the other one, we varied the \textit{number of ones}. 
\begin{itemize}
    \item \textbf{Varied number of bit split:} To generate the samples in this split, we first sampled the number of bits, then sampled each bit individually from a uniform Bernoulli distribution. For training, we used lengths between $3$ and $20$, and for validation/testing, we used lengths between $3$ and $40$. 
    
    \item \textbf{Varied number of ones split:} Here, we fixed the number of bits at $30$. To sample each instance, we first uniformly sampled the number of ones, then randomly placed each one in the fixed-length bitstring by randomly shuffling the bits. We used $10$ to $20$ ones in the training split, and $1$ to $30$ ones in the validation/test splits.
    
\textbf{Padded scratchpad:} The padded scratchpad format can be seen in \ref{fig:parity_instances}. Both the input and the scratchpad targets are padded left and right with the same number of padding tokens respectively, such that the relevant bit to attend to while constructing the scratchpad bits is always equidistant. Moreover the total number of characters/tokens is also kept constant. The number of tokens to pad on the left and right is determined (uniformly) randomly. 
\end{itemize}

The parity datasets contain 1000000 samples. 

\textbf{Natural Language Parity Dataset: } In order to tap into the natural language understanding capabilities of pretrained language models, we situated the parity task as a ``coin flip problem". In this framing, flipping a coin corresponds to $1$ and not flipping a coin corresponds to $0$. To make the inputs as close as possible to English without occupying too many tokens, we used the sentence templates ``Then <NAME> flips." and ``Then <NAME> doesn't flip." to represent whether the coin was flipped or not respectively, where ``<NAME>" refers to a randomly sampled given name. We also prepended each step with an integer id that count backwards from the total number of steps there are in the input sequence. We've experimented with versions where the integer ids are incremented. This didn't lead to a significant difference in the overall performance. 

\begin{figure}
\hspace*{-1.5cm}  
    \includegraphics[width=1.2
    \textwidth]{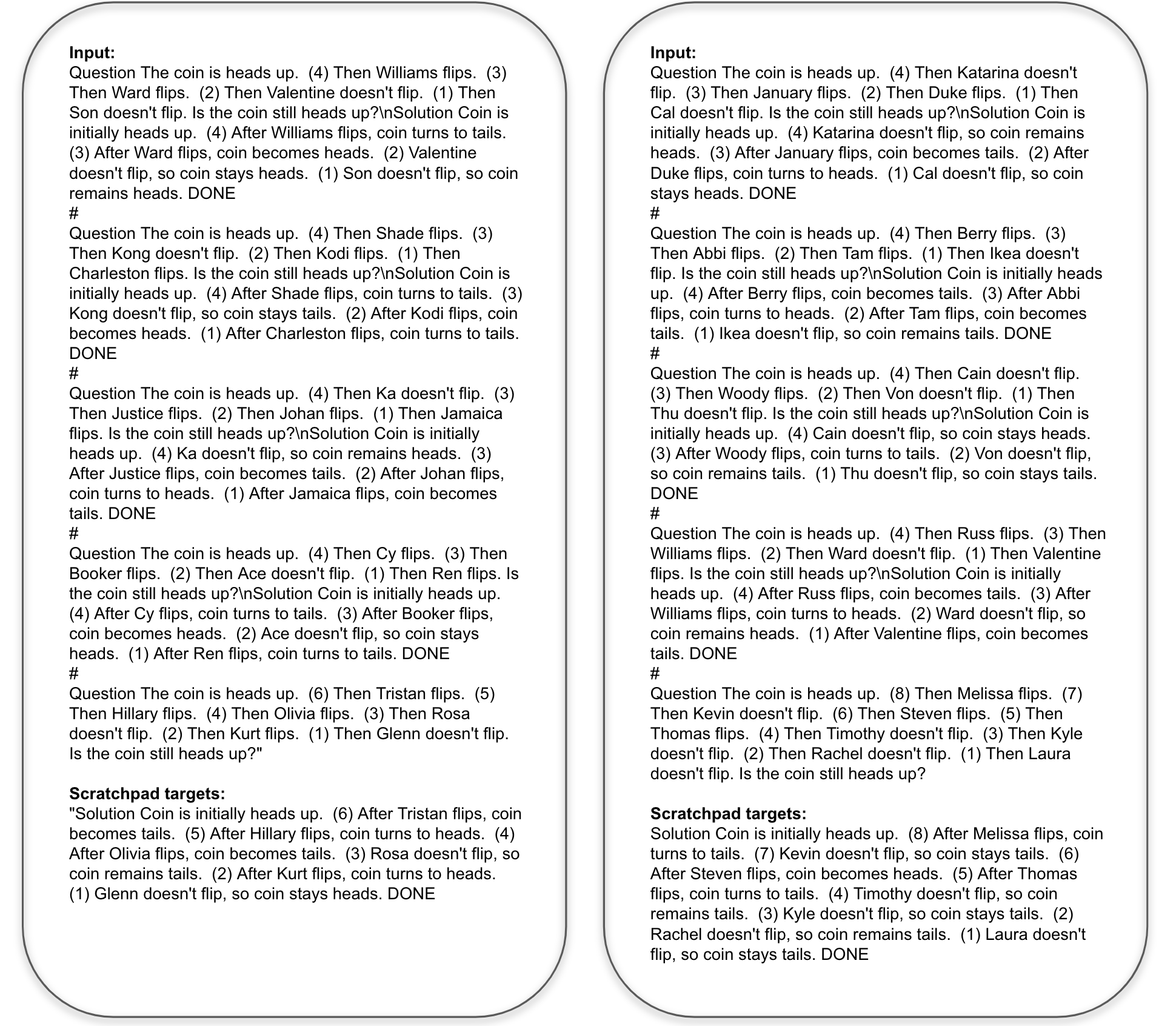}

    \caption{\small
    \textbf{Natural Language parity problem instances:} The coin flip task - which consists of tracking the state of a coin as it undergoes a number of flip or no-flip operations - shares the same underlying problem structure as the parity task. }
    \label{fig:parity_instances_natural}
\end{figure}

Two representative sample input-target pairs (including the exemplars) are provided in Figure \ref{fig:parity_instances_natural}.

\subsection{Boolean Variable Assignment Dataset:}

\begin{figure}
\hspace*{-1.5cm}  
    \includegraphics[width=1.2
    \textwidth]{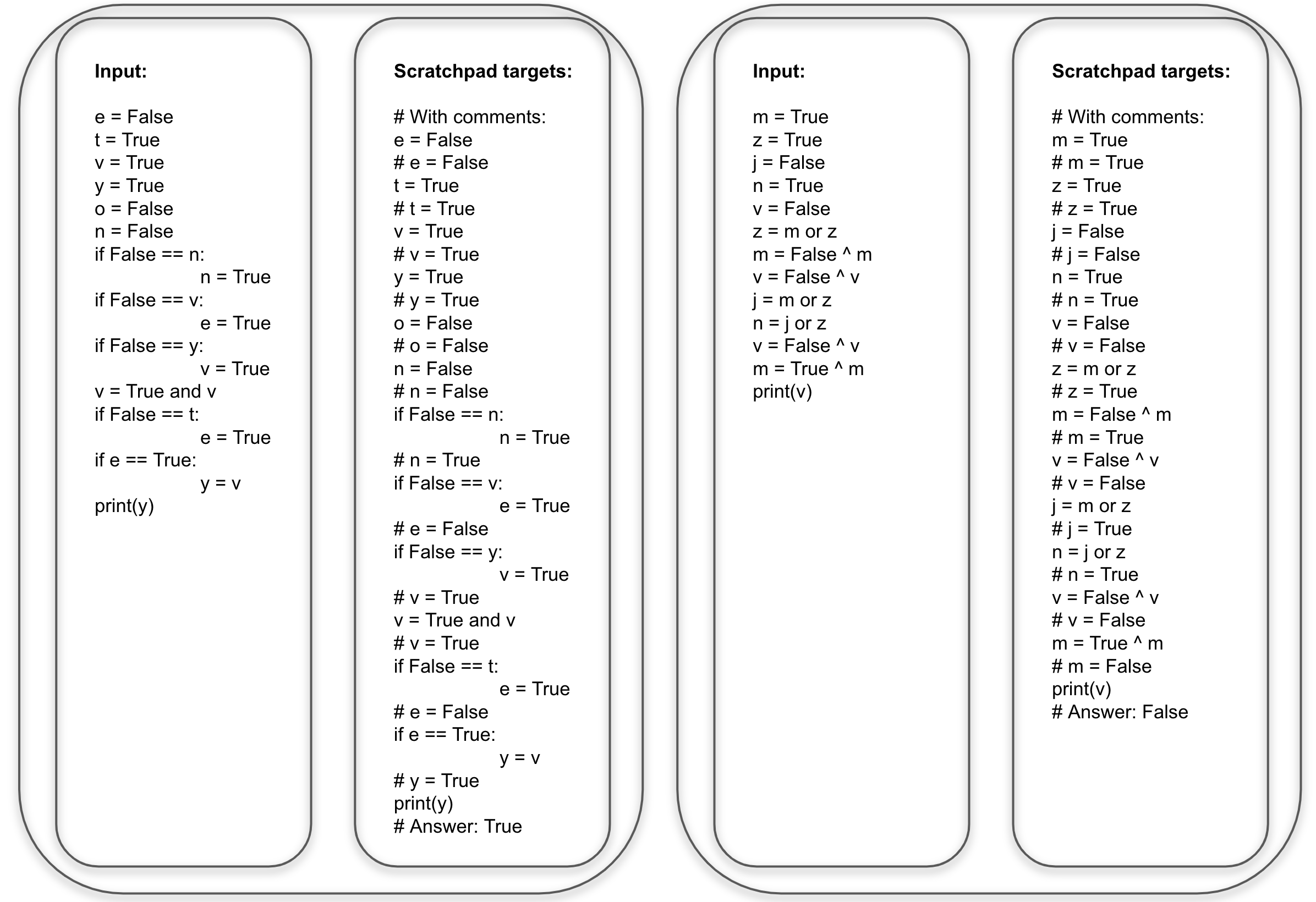}

    \caption{\small
    \textbf{Variable assignment problem instance:} A problem instance from the Boolean Variable Assignment dataset. The scratchpad strategy consists of outputting the value of the recenly updated variable in form of comments. Note that both the input and the scratchpad are valid Python programs.  }
    \label{fig:var_assign_instance}
\end{figure}

An instance of the variable assignment dataset can be seen in Figure \ref{fig:var_assign_instance}. The data generation procedure is aimed at synthesizing large number of qualitatively different programs: (1) A subset of boolean variable assignment operations is uniformly sampled from a large pool of operations. (2) The number of operations and number of variables (along with their names, which are single-letter characters) are uniformly sampled based on pre-set hyperparameters. (3) One by one, operations and the variables included in the operations are sampled, while making sure that each added operations retains the semantic correctness of the program. 

We now outline the hyperparameters used to generate the \textit{chain-like} and \textit{diverse} splits. 

\textbf{Chain-like split:} 
\begin{itemize}
    \setlength\itemsep{1mm}
    \item \textbf{Boolean operators:} assign to \textit{and} with another variable, assign to \textit{or} with another variable, assign to \textit{xor} with another variable, negate
    \item \textbf{Minimum/maximum number of operations:} 3, 19
    \item \textbf{Minimum/maximum number of variables in the program:} 2, 3
\end{itemize}

\textbf{Diverse split:} 
\begin{itemize}
    \setlength\itemsep{1mm}
    \item \textbf{Boolean operators:} assign to \textit{and} with another variable, assign to \textit{or} with another variable, assign to \textit{xor} with another variable, negate, assign to \textit{and} with boolean, assign to \textit{or} with boolean, assign to \textit{xor} with a boolean, conditional assign to a boolean, assign to another variable, conditional assign to another variable
    \item \textbf{Minimum/maximum number of operations:} 8, 32
    \item \textbf{Minimum/maximum number of variables in the program:} 4, 10
\end{itemize}

The scratchpad format (seen in Figure \ref{fig:var_assign_instance}) consists of copying over the input program with comments after each line specifying the value of the recently updated variable. This ensures that the scratchpad itself is a valid Python program and can be used with models pretrained with Python data. 

Both splits have 1500000 samples.

\section{Baseline for Vanilla Finetuning on Variable Assignment}
Just how weak are the vanilla finetuned models on the OOD lengths on the variable assignment task? We trained baseline models with the same parameter count on a modified version of the variable assignment dataset where the order of the operations were randomly shuffled. While this leaves in some of the spurious features that correlate with the right answer, it completely eliminates the possibility of running a sequential algorithm to get to the final answer. 

\begin{figure}
\centering
    \includegraphics[width=0.7
    \textwidth]{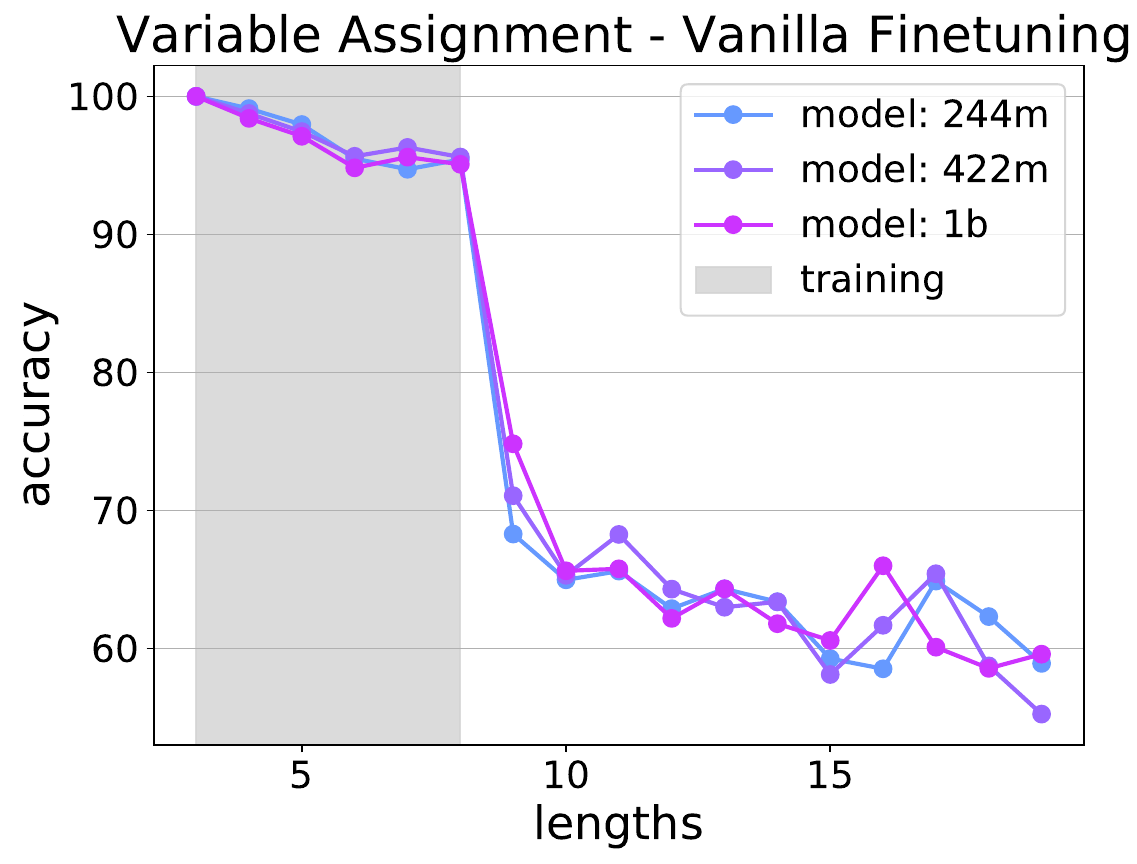}

    \caption{\small
    \textbf{Results of finetuning on the shuffled-ops variable assignment dataset:} We trained baseline models with the same parameter counts on a modified version of the variable assignment dataset where the order of the operations were randomly shuffled.
    The generalization gap between in and out-of-distribution data persists here as well, due to transformers' tendency to prefer parallel strategies that don't generalize to larger lengths. }
    \label{fig:shuffled_ops_baseline}
\end{figure}

The results can be found in Figure \ref{fig:shuffled_ops_baseline}. On OOD data, the performance of the shuffled-ops baseline is on par with the models trained with the clean version of the dataset. Note that the length generalization deficiency that the shuffled ops baselines display is even more dramatic than that of the models' trained with clean data. This is not surprising, as the primary source of lack of length generalization is the transformers' tendency to prefer parallel strategies that don't generalize to larger lengths over picking up sequential algorithm

\section{Experimental Conditions}shuff
We outline the training conditions and hyperparameter used in the main experiments. 

In our experiments on different datasets, we first did a learning rate sweep over different model sizes, and preferred the largest learning rates that ensured training stability. This is due to the fact that we observed the best length generalization when we used larger learning rates (see Figure \ref{fig:lg-subtleties}). We used the AdaFactor optimizer in all of our finetuning experiments~\citep{shazeer2018adafactor}. We didn't use dropout~\citep{srivastava2014dropout} in the parity experiments and used a dropout rate of $0.05$ in the variable assignment experiments. Our initial experiments suggest that one can often reach similar in and out-of-distribution performance either by training the weights from scratch, or finetuning from the pretrained weights. To keep the experiments consistent, we always initialized training using the pretrained weights. In variable assignment, we did a learning rate sweep over $0.0033$, $0.00033$ and $0.000033$. For parity, we search over learning rate velus of $0.002$, $0.0002$ and $0.00002$. We used a constant learning rate profile all throughout learning. Due to memory constraints, we used a batch size of $32$ when training the $64b$ and $128b$ models. 

We used greedy decoding in all of our experiments (including few-shot scratchpad ones). We experimented with temperature sampling with reranking based on sentence likelihoods, but found that doing this doesn't lead to qualitatively different results.

\section{Computational Graph Depth is the Relevant Notion of Difficulty on Variable Assignment}
\begin{figure}
    \includegraphics[width=1.
    \textwidth]{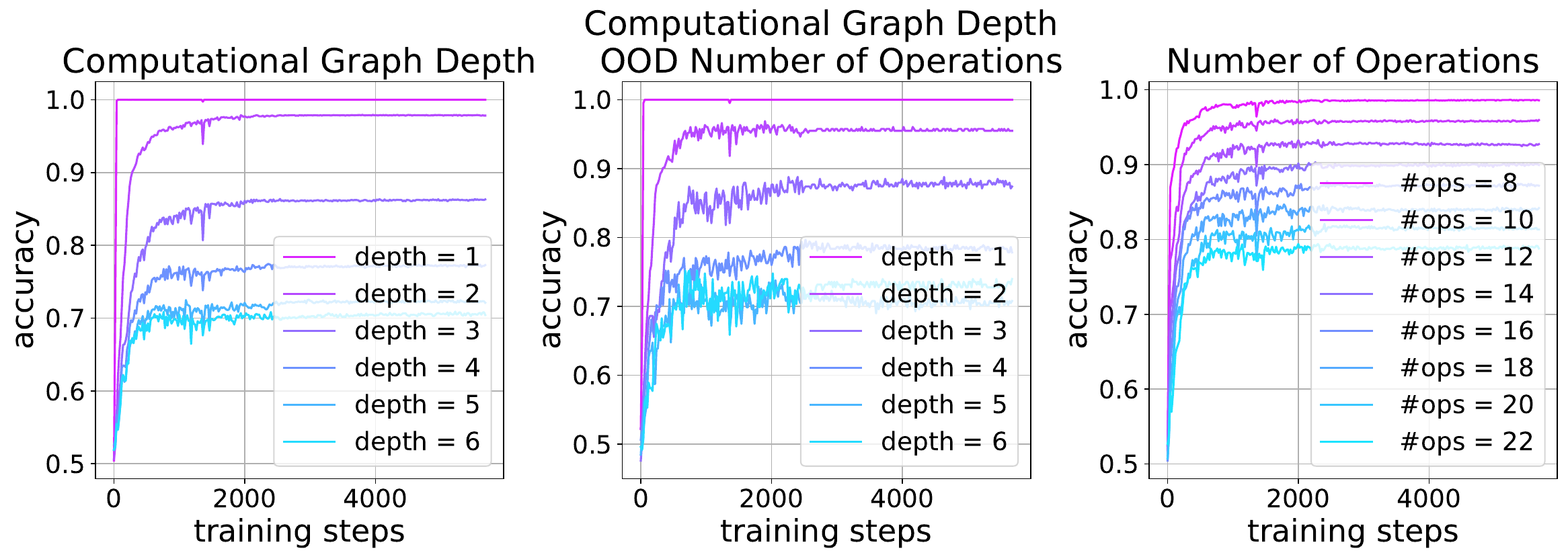}

    \caption{\small
    \textbf{Evolution of Performance over Training Iterations on Different Computational Graph Depths:} Computational graph depth corresponds to the length of the longest dependency chain linking to the queried variable in the variable assignment task. This quantity captures a more suitable notion of length/difficulty for transformer models. On the \textbf{left} plot, we show how the accuracy of a transformer model evolves for problem instances with different computational graph depths. In the \textit{middle}, we show the same, except that we constrain the number of operations in the program to an out-of-distribution number. The accuracy values roughly remain unchanged, indicating that it's not the number of operations, but computational graph depth that determines the difficulty of a problems instance. On the right, we show the evolution of accuracy on instances with different number of operations for reference. }
    \label{fig:comp_graph_depth_fig}
\end{figure}

\textit{Computational graph depth} captures a more relevant notion of difficulty on the variable assignment task for transformers. In Figure \ref{fig:comp_graph_depth_fig}, we show how the accuracy of a transformer model evolves on samples of problem instances with different computational graph depths (left), and how the same quantity behaves if we fix the number of operations in a program at an out-of-distribution length (middle)~\footnote{The longest in-distribution number of operations is $15$, and we fix this quantity at $20$ in the middle plot.}. Two takeaways from this analysis are:
\begin{itemize}
    \item The fact that the accuracy values on samples with different computational graph depths roughly remain unchanged on OOD program lengths indicates that it's not the number of operations, but computational graph depth that captures a more relevant notion of difficulty. 
    \item Transformers initially pick up how to handle programs with a small computational graph depth throughout training and then move to more difficult programs. 
\end{itemize}

\section{Effect of Prompt Style on Few-Shot Finetuning Performance}
\begin{figure}
    \includegraphics[width=1.
    \textwidth]{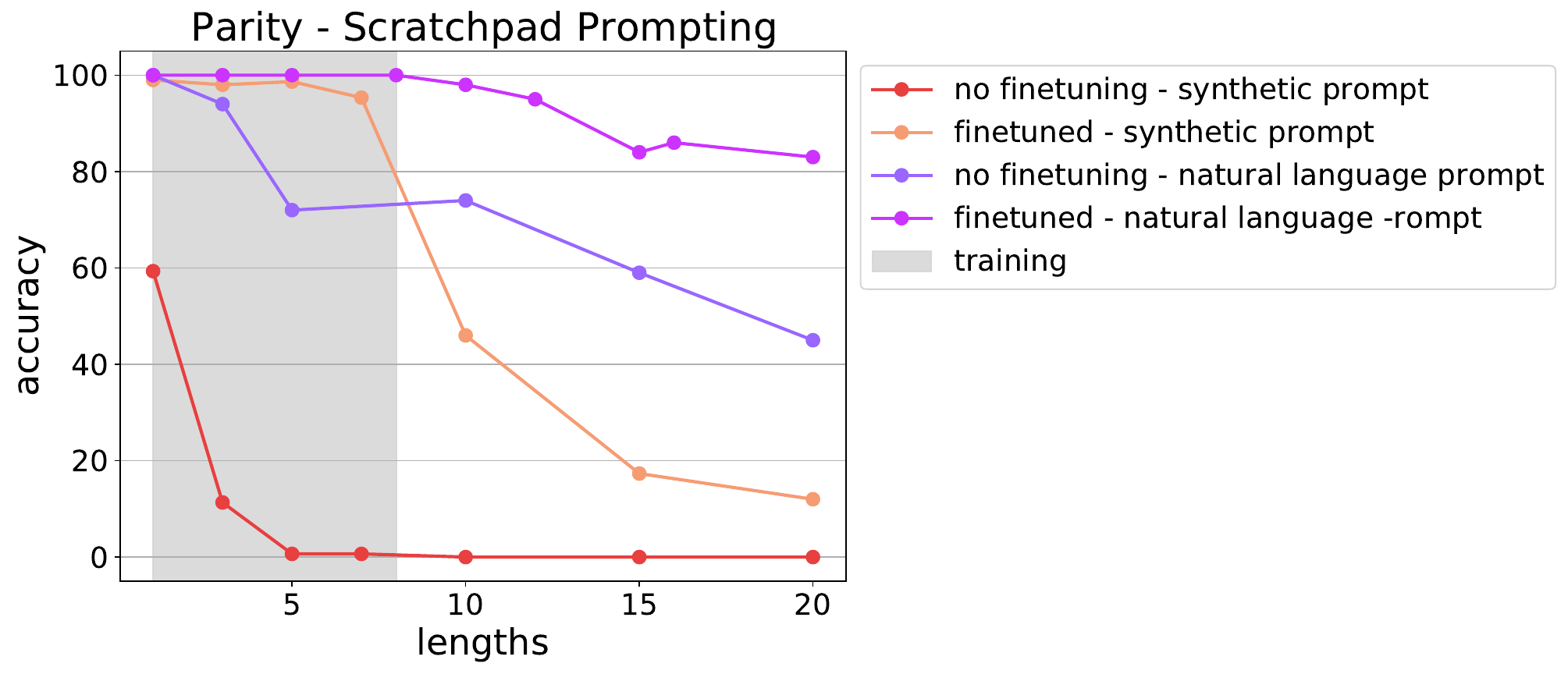}

    \caption{\small
    \textbf{Effect of prompt style on few-shot finetuning performance:} we evaluated the (few-shot) finetuned performance of the pretrained model on an alternative, synthetic prompt style that yields poor performance without any pretraining. We observed that the few-shot finetuned model on this task also shows significant length generalization pathologies. This is in line with our hypothesis that the non-finetuned performance of the base model should be non-trivial for few-shot finetuning to consistently yield strong length generalization results. }
    \label{fig:effect_of_prompt_style}
\end{figure}

In Section \ref{sec:few-shot-sp}, we hypothesized that few-shot finetuning only leads to significant improvements in length generalization performance if the non-finetuned performance on the same task already at a nontrivial level. To provide a sanity check for this, we ran few-shot finetuning using an alternative prompt style for the coin-flip task that yields poor non-finetuned performance. As can be seen in Figure \ref{fig:effect_of_prompt_style}, the few-shot finetuned performance shows significant length generalization pathologies. We leave a systematic study of how prompt style affects length generalization as future work.

\section{Distractor Analysis for Scratchpad Strategies}
Our analysis in Section \ref{sec:scratchpad_finetuning} indicates that length generalization pathologies persist even when we use the padded scratchpad strategy that makes sure that it's not untrained position encodings and/or the EOS token prediction that causes the aforementioned pathologies. This points to the fact that the transformer doesn't learn to attend to the ``right" section of the input and scratchpad that implements the sequential strategy that generalizes to longer lengths --- it's thrown off by distractor tokens in the input and/or the preceding scratchpad targets.
The distractor tokens at which section of the transformer context window (input or scratchpad) contribute more to the performance deterioration? If we remove all the distractor tokens, can we achieve perfect length generalization? 

To answer these questions, we trained four transformer models under the following conditions: (1) We used the padded scratchpad strategy described in Section \ref{sec:scratchpad_finetuning} with no modification, (2) We manually masked the preceding scratchpad tokens that don't contribute to the correct sequential algorithm (i.e. we masked the distractor tokens in the input), (3) We masked the input tokens that don't contribute to the correct sequential algorithm, (4) we masked the distractor tokens in both the input and the preceding distractor tokens. To give an example, let's say the input bitstring is ``[1 1 0 1 1]", and the model has emitted the scratchpad tokens ``[1 0 0]" so far. Masking the distracting sratchpad tokens simply means replacing all but the last scratchpad token with dummy padding tokens: ``[x x 0]". Similarly, removing the distracting input tokens corresponds to masking out the part of the input that the network doesn't need to attend to while predicting the next bit: ``[x x x 1 x]". Masking both corresponds to removing the distractor tokens in both the input and the target, so that the input and the preceding scratchpad become ``[x x x 1 x]" and ``[x x 0]" respectively. 

\begin{figure}
    \includegraphics[width=1.
    \textwidth]{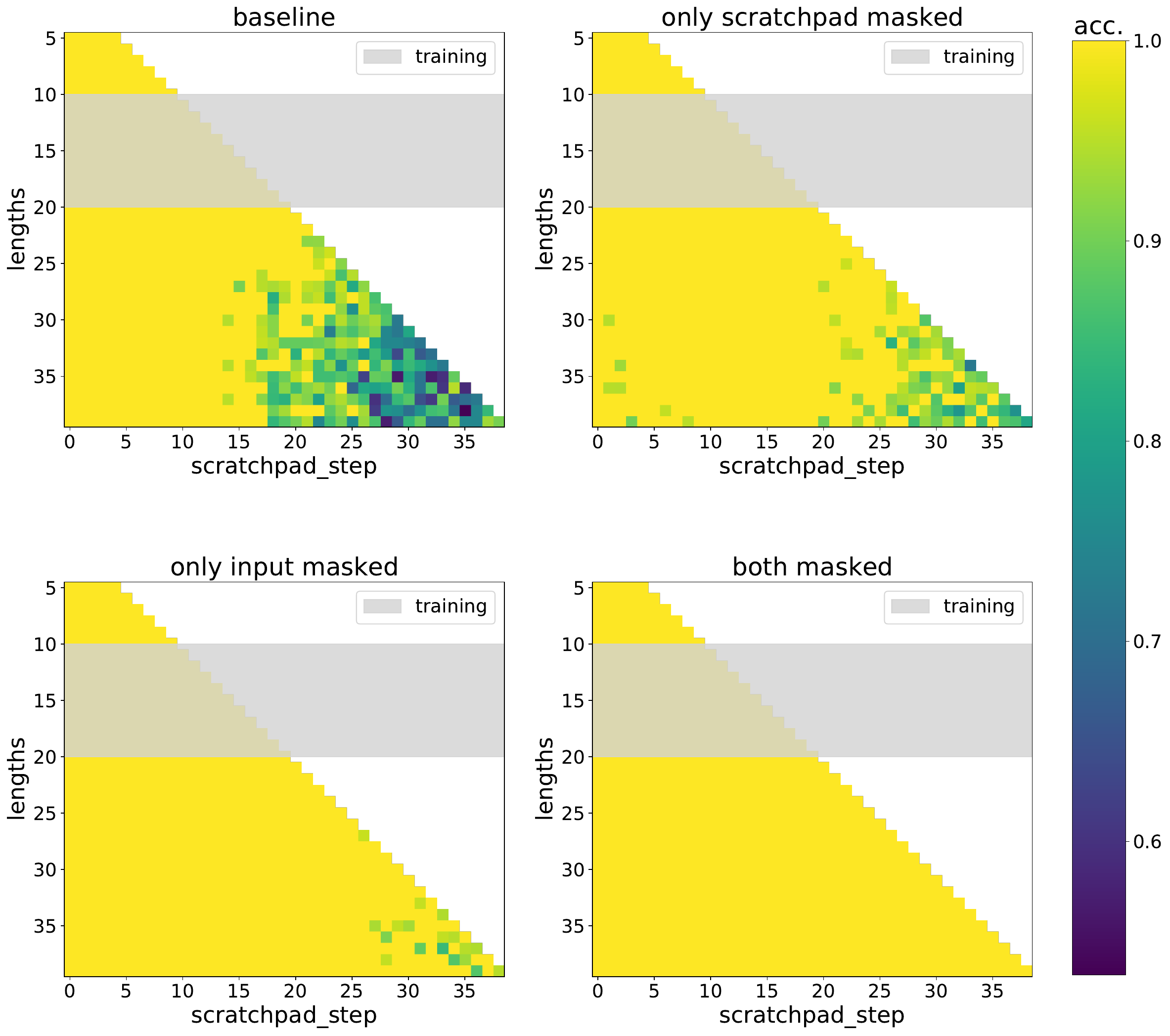}

    \caption{\small
    \textbf{Distractor analysis:} We trained scratchpad-augmented transformers to solve the parity task, where we systematically masked out the tokens in the input (left bottom) and the scratchpad (right top) that don't need to be attended to while implementing the correct sequential algorithm that solves parity. The plots illustrate the accuracy of the trained models on predicting the scratchpad tokens at different steps for inputs of varying (in and OOD) lengths. This analysis shows that (1) removing all distracting tokens leads to perfect length generalization (right bottom), (2) the distractor tokens in the input contribute more to the length generalization pathologies (right top versus left bottom).  }
    \label{fig:distractor_analysis}
\end{figure}

The results can be seen in Figure \ref{fig:distractor_analysis}. For all four experimental conditions, we plotted the accuracy of the trained models on predicting the scratchpad tokens at different steps for inputs of varying (in and OOD) lengths. We conclude from this experiment that:
\begin{itemize}
    \item Removing all distractor tokens does result in perfect length generalization. 
    \item The distracting input tokens are hurt length generalization performance more. 
\end{itemize}
Based on this analysis, we conclude that innovations in transformer architectures and/or training methodology/objective that alleviate the issues caused by distractor tokens have a chance at significantly improving length generalization.

\end{document}

%% file: macros.tex
\newcommand{\cmark}{\ding{51}}%
\newcommand{\xmark}{\ding{55}}%
\newcommand{\pretrainedmodel}{\texttt{LaMDA}~}%
\newcommand{\efficacyhigh}{\textcolor{green}{\cmark\cmark}}
\newcommand{\efficacymed}{\textcolor{orange}{\cmark}}
\newcommand{\efficacylow}{\textcolor{red}{\xmark}}